\documentclass[final]{clv2025}

\jvol{vv}
\jnum{nn}
\jyear{2025}
\dochead{Short Paper}
\usepackage{amsmath}
\usepackage{booktabs}
\usepackage{cleveref}
\usepackage{multirow}
\usepackage{xspace}
\usepackage{dsfont}
\usepackage{siunitx}
\usepackage{pifont}

\newcommand{\pojsd}{PO-JSD\xspace}
\newcommand{\metric}[1]{\textbf{#1}\xspace}
\newcommand{\eqdef}{\overset{\mathrm{def}}{=}}  % equal sign meaning 'is defined as'
\newcommand{\methname}[1]{\textbf{#1}}  % format method name e.g. ReL
\newcommand{\yes}{\textcolor{green}{\ding{51}}\xspace}
\newcommand{\no}{\textcolor{red}{\ding{55}}\xspace}

\DeclareMathOperator{\mgt}{>\!\!>}
\DeclareMathOperator{\JSD}{JSD}
\DeclareMathOperator{\KL}{KL}

\runningtitle{Training and Evaluating with Human Label Variation}
\runningauthor{Kurniawan, Mistica, Baldwin, Lau}

\begin{document}

\title{Training and Evaluating with Human Label Variation: An Empirical Study}

\author{Kemal Kurniawan\thanks{Corresponding author}$^{,1}$, Meladel
  Mistica$^{1}$, Timothy Baldwin$^{2,1}$, Jey Han Lau$^{1}$}

\affilblock{
    \affil{School of Computing and Information Systems, University of Melbourne\\\quad \email{kurniawan.k@unimelb.edu.au}}
    \affil{Mohamed bin Zayed University of Artificial Intelligence}
}

\maketitle

\begin{abstract}
  Human label variation~(HLV) challenges the standard assumption that
  a labelled instance has a single ground truth, instead embracing the natural
  variation in human annotation to train and evaluate models. While
  various training methods and metrics for HLV have been proposed,
  it is still unclear which methods and metrics perform best in what settings.
  We propose new evaluation metrics for HLV leveraging fuzzy set theory.
  Since these new proposed metrics are differentiable, we then in turn
  experiment with employing these metrics as training objectives. We conduct an
  extensive study over 6 HLV datasets testing 14 training methods and 6
  evaluation metrics. We find that training on either disaggregated annotations
  or soft labels performs best across metrics, outperforming training using the
  proposed training objectives with differentiable metrics. We also show that
  our proposed soft micro F\textsubscript{1} score is one of the best metrics
  for HLV data.\footnote{Code and supplementary materials are available at:
    \url{https://github.com/kmkurn/train-eval-hlv}}
\end{abstract}

\section{Introduction}

Human label variation~(HLV) challenges the standard assumption that an
example has a single ground truth~\citep{plank2022}. With HLV, different
human annotations for the same instance are viewed as useful signal rather than
undesirable noise. In this context, model training and evaluation are no longer straightforward. For
example, standard metrics such as accuracy assume that a single ground truth label
exists for each instance. To address this issue, recent
work~\citep[\textit{inter
  alia}]{pavlick2019,peterson2019,nie2020,uma2020,uma2021,cui2023,gajewska2023,maity2023,wan2023}
has proposed various methods to train models and metrics to evaluate their
performance in the HLV context. However, the lack of a systematic study means it
is still unclear which methods and metrics perform best.

In natural language processing (NLP), most existing metrics for HLV represent human judgements as probability
distributions over classes and employ information-theoretic measures such as
divergence. It has largely ignored the field of remote sensing that has dealt
with issues in model evaluation against fuzzy ground
truths~\citep{foody1996,binaghi1999,lewis2001,pontius2006,pontius2006a,silvan-cardenas2008,gomez2008}.
In remote sensing, the task is to classify patches of a satellite image of
a region into categories such as land, water, and so forth. As an image
patch often contains multiple categories, the field has developed evaluation
metrics that can handle this variation in the ground truth labels.

Taking inspiration from remote sensing research, we represent human
judgement distributions as degrees of membership over fuzzy sets and generalise
standard metrics such as accuracy into their soft versions using fuzzy set
operations. In contrast to information-theoretic measures, each of these soft
metrics has an intuitive interpretation that corresponds to the standard
counterpart. For example, soft accuracy is interpreted as the proportion of
correctly predicted judgement distributions. To the best of our knowledge, we
are the first to propose employing soft metrics for HLV in NLP.

As these soft metrics are differentiable, we next explore the use of each soft
metric as the training objective. We perform an extensive study testing 14
training methods and 6 evaluation metrics on 6 HLV datasets across 2 pretrained
models of different sizes to investigate which training methods are best for HLV
data across the different metrics.
This study is then followed by an empirical meta-evaluation of the evaluation
metrics to understand which evaluation metrics are best for HLV data. To the
best of our knowledge, we are the first to perform such an empirical
meta-evaluation of HLV evaluation metrics.

To summarise, we: (a)~propose soft metrics to measure model performance in the
context of HLV, taking inspiration from remote sensing research; (b)~propose new
training methods based on the soft evaluation metrics; and (c)~conduct an
extensive study covering 6 HLV datasets, 14 training methods, 2 pretrained
models, and 6 evaluation metrics to understand the best training methods and
evaluation metrics.

We find that two of the simplest methods for HLV training perform surprisingly
well, outperforming more complex methods including training with each
soft metric as the objective and attaining the best performance in most cases. The
first method considers each annotation of an instance as a separate
instance--label pair. Training is then performed on this disaggregated data where
the same instance may occur multiple times with a different label each. The
second method trains the model to predict the full label distribution induced by
the instance's annotations using the standard cross-entropy loss. This is in
contrast to the standard model training where the prediction target is the mode
of the label distribution.

For evaluation metrics, we find that an existing metric based on Jensen-Shannon
divergence~(JSD) and our proposed soft micro F\textsubscript{1} score are two of the
best HLV metrics. Our analysis shows that in single-label classification, the
two metrics are highly correlated but we prove that soft accuracy~(because micro
F\textsubscript{1} is equal to accuracy in this case) is upper-bounded by the
JSD-based metric. As a result, we show that the JSD-based metric can give a
score close to its maximum value to wrong predictions, which is potentially
misleading, unlike our proposed soft accuracy. We thus recommend future work in
HLV to report both metrics but focus on our proposed soft metrics when
interpretability is important.

\section{Related Work}\label{sec:related-work}

\paragraph{Training methods for human label variation}

First introduced by \citet{plank2022}, the term \textit{human label
  variation}~(HLV) captures the fact that disagreeements in annotations can be
well-founded and thus signal for data-driven methods. It challenges the
traditional notion in machine learning that an instance has a single ground
truth. Training methods that accommodate such variation had been proposed before
the term was coined~\citep[\textit{inter alia}]{sheng2008,peterson2019,uma2020}.
Newer methods have also been proposed in the
literature~\citep{deng2023,lee2023,chen2024,rodriguez-barroso2024} and in the
Learning with Disagreement shared task~\citep{leonardelli2023} which provides a
benchmark for HLV. While a systematic investigation of some of these methods
exists~\citep{uma2020}, it employs smaller pretrained language models and covers
only binary or multiclass classification tasks. In contrast, our work
additionally employs a large language model and covers multilabel tasks.
Furthermore, we also propose new soft evaluation metrics for HLV.

\paragraph{Evaluation of soft classification in remote sensing}

While evaluating in the HLV context is a relatively new concept in NLP,
evaluating against a fuzzy\footnote{The terms ``fuzzy'' and ``crisp'' are sometimes
  used in remote sensing literature to mean ``soft'' and ``hard'' in NLP
  literature.} reference is a well-studied area in remote sensing research.
Early work by \citet{foody1996} used a measure of distance that is equivalent to
twice the Jensen-Shannon divergence~\citep{lin1991}. \citet{binaghi1999} argued
that entropy-based measures are sensible only if the reference is crisp, and
proposed the use of fuzzy set theory to compute the soft version of standard
evaluation metrics such as accuracy. A similar approach was also proposed by
\citet{harju2023} for audio data. Key to this approach is the minimum function
used to compute the class membership scores given a fuzzy reference and a fuzzy
model output. Other proposed functions include product~\citep{lewis2001},
sum~\citep{pontius2006a}, and a composite of such functions designed to ensure
diagonality of the confusion matrix when the reference and the model output
match perfectly~\citep{pontius2006,silvan-cardenas2008}. Approaches that take
the cost of misclassifications into account have also been
proposed~\citep{gomez2008}. In this work, we take the fuzzy set approach by
\citet{binaghi1999} and apply it to text classification tasks.

\paragraph{Meta-evaluation of HLV evaluation metrics}

There is little existing work on the meta-evaluation of HLV metrics. Most
studies used information theory-based measures such as
cross-entropy~\citep[\textit{inter alia}]{uma2020,leonardelli2023}, presumably
because of the standard probabilistic approach in modern NLP. \citet{rizzi2024}
proposed some theoretical properties that an ideal soft metric should satisfy
and performed a theoretical meta-evaluation of existing HLV metrics for
single-label classification. In contrast, we propose new soft metrics for both
single- and multilabel classification and perform an \emph{empirical}
meta-evaluation of these metrics.

\section{Evaluation Metrics for HLV}\label{sec:eval-metrics}

We now discuss the evaluation metrics relevant for HLV data, first on
multiclass metrics (\Cref{sec:metrics}), then multilabel metrics
(\Cref{sec:mullab-metrics}), and finally the relationship between these metrics
(\Cref{sec:rel-overall-metrics}). In these discussions, we will also introduce our use of
soft metrics and explain how they relate to their hard variant counterparts.
\Cref{tbl:metrics} provides a summary of the existing and proposed/new
evaluation metrics.

\paragraph{Notation}

Let $P_{ik}\geq 0$ denote the human judgement for example $i$ and class
$k$, i.e.\ proportion of humans labelling example $i$ with class $k$.
Let $Q_{ik}\geq 0$ denote the value of $P_{ik}$
predicted by a model.
An evaluation metric $m$ is a function such that the scalar
$m(\mathbf{P},\mathbf{Q})$ measures how well~(i.e., a positive orientation)
$\mathbf{Q}$ captures the judgement distribution in $\mathbf{P}$.
To complete our
notation, let $N$ and $K$ denote the number of examples and classes
respectively.

\begin{table}
  \caption{HLV metrics for classification tasks.}\label{tbl:metrics}
  \begin{tabular}{@{}lll@{}}
    \toprule
                               & Multiclass                    & Multilabel                     \\
    \midrule
    \multirow{4}{*}{Existing:} & Accuracy                      & Micro F\textsubscript{1}       \\
                               & Macro F\textsubscript{1}      & Macro F\textsubscript{1}       \\
                               & \pojsd                        &                                \\
                               & Entropy correlation           &                                \\
    \cmidrule{1-3}
    \multirow{4}{*}{Proposed:} & Soft accuracy                 & Soft micro F\textsubscript{1}  \\
                               & Soft macro F\textsubscript{1} & Soft macro F\textsubscript{1}  \\
                               &                               & Multilabel \pojsd              \\
                               &                               & Multilabel entropy correlation \\
    \bottomrule
  \end{tabular}
\end{table}

\subsection{Multiclass Metrics}\label{sec:metrics}

In multiclass classification tasks, an example can only be assigned to one
class. This is reflected by the unity constraints $\sum_{k}P_{ik}=1$ and
$\sum_{k}Q_{ik}=1$ for all $i$.  We study a total of 6 multiclass evaluation
metrics $m$. The first two are standard hard metrics, the next two are their
soft versions that we propose inspired by prior
work~\citep{binaghi1999,harju2023}, and the last two are existing HLV metrics.

Let $I^P_{ik}$ and $I^Q_{ik}$ denote $\mathds{1}(\arg\max_{l}P_{il}=k)$ and
$\mathds{1}(\arg\max_{l}Q_{il}=k)$ respectively. The 6 metric definitions under our
notation are as follows:
\begin{enumerate}
  \item \metric{(Hard)~accuracy}, where:
    \begin{equation*}
      m(\mathbf{P},\mathbf{Q})\eqdef\frac{1}{N}\sum_{ik}I^P_{ik}I^Q_{ik}.
    \end{equation*}
    This is the standard evaluation metric for classification tasks.
  \item \metric{(Hard)~macro F\textsubscript{1}}, where
    \begin{equation*}
      m(\mathbf{P},\mathbf{Q})\eqdef\frac{1}{K}\sum_{k}\frac{2\sum_{i}I^P_{ik}I^Q_{ik}}{\sum_{i}\left(I^P_{ik}+I^Q_{ik}\right)}.
    \end{equation*}
    This metric addresses the issue with accuracy
    that is biased toward the majority class when class proportions are
    imbalanced.
  \item \metric{Soft accuracy}, where:
    \begin{equation*}
      m(\mathbf{P},\mathbf{Q})\eqdef\frac{1}{N}\sum_{ik}\min(P_{ik},Q_{ik}).
    \end{equation*}
    This metric is novel to this work, for evaluation in the context of HLV.
    This is a special case of the soft micro F\textsubscript{1} score proposed
    later in \Cref{sec:mullab-metrics}.
  \item \metric{Soft macro F\textsubscript{1}}, where:
    \begin{equation*}
      m(\mathbf{P},\mathbf{Q})\eqdef\frac{1}{K}\sum_{k}\frac{2\sum_{i}\min(P_{ik},Q_{ik})}{\sum_{i}(P_{ik}+Q_{ik})}.
    \end{equation*}
    This metric is also novel to this work, to address the issue of class imbalance with
    soft accuracy and other existing HLV metrics.\footnote{In the
      context of HLV, we understand the concept of class imbalance
      analogously to that in the context where class assignments are
      hard: there exists a class $k$ such that
      $\sum_{i}P_{ik}\mgt\sum_{i}P_{il}$ for all $l\neq k$.
    } Note that this metric implies the existence of soft
    \emph{class-wise} metrics~(see Appendix).
  \item \metric{Jensen-Shannon divergence}~\citep{uma2021}, which we modify into its
    positively oriented version~(\metric{\pojsd} for short) where:
    \begin{equation*}
      m(\mathbf{P},\mathbf{Q})\eqdef1-\frac{1}{N}\sum_{i}\JSD(\mathbf{p}_i,\mathbf{q}_i).
    \end{equation*}
    Vectors $\mathbf{p}_i,\mathbf{q}_i$ denote the $i$-th row of
    $\mathbf{P},\mathbf{Q}$ respectively. The scalar
    $\JSD(\mathbf{a},\mathbf{b})=\frac{1}{2}\left(\KL(\mathbf{a},\frac{1}{2}\left(\mathbf{a}+\mathbf{b}\right))+\KL(\mathbf{b},\frac{1}{2}\left(\mathbf{a}+\mathbf{b}\right))\right)$
    is the Jensen-Shannon divergence~\citep{lin1991}, where
    $\KL(\mathbf{a},\mathbf{b})=\sum_{k}a_k\log_2\frac{a_k}{b_k}$ is the
    Kullback-Leibler divergence.\footnote{Jensen-Shannon divergence of two
      distributions has an upper bound of $\log_b2$ if the logarithms used in
      $\KL(\cdot)$ are of base $b$. Normalising this bound to 1 results in
      logarithms of base 2 as $\log_b x/\log_b2=\log_2x$.}
  \item \metric{Entropy correlation}~\citep{uma2020}, where:
    \begin{align*}
      m(\mathbf{P},\mathbf{Q})&\eqdef\frac{\sum_i\zeta^P_i\zeta^Q_i}{\sqrt{\sum_i{\zeta^P_i}^2}\sqrt{\sum_i{\zeta^Q_i}^2}},\\
      \zeta^P_i&=\eta^P_i-\bar{\eta}^P,\text{ and }\\
      \zeta^Q_i&=\eta^Q_i-\bar{\eta}^Q.
    \end{align*}
    Scalar $\eta^P_i=-\sum_k\frac{P_{ik}\log P_{ik}}{\log K}$ is the normalised
    entropy of row $i$ of $\mathbf{P}$, and
    $\bar{\eta}^P=\frac{1}{N}\sum_{i}\eta^P_i$ is the mean of $\lbrace
    \eta^P_i\rbrace_{i=1}^N$. Scalars $\eta^Q_i$ and $\bar{\eta}^Q$ are defined
    analogously.
\end{enumerate}

\paragraph{Interpretation of soft accuracy}

By noting that $N=\sum_{ik}P_{ik}$, soft accuracy can be interpreted as the
proportion of the judgement distribution that is correctly predicted. For
example, if soft accuracy is 70\%, then the model predicts 70\% of judgement
distribution correctly. This interpretation is similar to the hard
counterpart~(i.e.\ proportion of examples that are correctly predicted) and
arguably more intuitive than that of existing metrics because it is stated
directly in terms of the problem of predicting human judgement distribution.

Besides interpretation, soft accuracy's similarities to the hard counterpart
include symmetry~(due to the symmetry of $\min$) and boundedness~(between 0 and
1), where the lower and the upper bounds are achieved if and only if
$P_{ik}Q_{ik}=0$ and $P_{ik}=Q_{ik}$ respectively for all $i,k$. Note also that
\emph{soft accuracy is reduced to the standard hard accuracy} when all rows of
both $\mathbf{P}$ and $\mathbf{Q}$ are one-hot vectors~(i.e.\ no label
variation).

While functions other than $\min$ have been used to compute soft accuracy~(see
\Cref{sec:related-work}), they lead to the soft accuracy being $<1$ in the
case where $\mathbf{P}=\mathbf{Q}$. This is problematic because intuitively,
good HLV evaluation metrics should produce the highest score when the predicted
judgement distribution perfectly matches the true counterpart. Therefore, we
use the $\min$ function in this work.

Other evaluation metrics we considered included entropy similarity~\citep{uma2021},
(negative)~cross-entropy~\citep{peterson2019,pavlick2019},
(negative)~Kullback-Leibler divergence~\citep{nie2020}, (the positively oriented
version of)~information closeness~\citep{foody1996} and (the positively oriented
version of)~Jensen-Shannon distance~\citep{nie2020}. However, we ultimately excluded them
from this study because of their expected high correlation with either \pojsd
or entropy correlation.

\subsection{Multilabel Metrics}\label{sec:mullab-metrics}

Next, we consider the case where the unity constraints do not hold,
i.e.\ an example can be (soft)~assigned to multiple classes, forming a
\emph{multilabel} classification task.

By considering this case as multiple independent binary classification tasks,
some existing metrics can be extended straightforwardly by taking the average of
the metric values over classes. For example, \pojsd can be modified into the
multilabel version:
\begin{equation*}
  m(\mathbf{P},\mathbf{Q})\eqdef1-\frac{1}{NK}\sum_{ik}\JSD(P^{\text{bin}}_{ik},Q^{\text{bin}}_{ik})
\end{equation*}
where $P^{\text{bin}}_{ik}=[P_{ik},1-P_{ik}]$ is a judgement distribution over
two classes, and $Q^{\text{bin}}_{ik}$ is defined analogously.
Entropy correlation can also be modified similarly into the multilabel version:
\begin{equation*}
  m(\mathbf{P},\mathbf{Q})\eqdef\frac{1}{K}\sum_{k}\frac{\sum_i\zeta^P_{ik}\zeta^Q_{ik}}{\sqrt{\sum_i{\zeta^P_{ik}}^2}\sqrt{\sum_i{\zeta^Q_{ik}}^2}}
\end{equation*}
where $\zeta^P_{ik}=\eta^P_{ik}-\bar{\eta}^P_k$,
$\eta^P_{ik}=-\frac{1}{\log 2}\left( P_{ik}\log P_{ik}+(1-P_{ik})\log(1-P_{ik})
\right)$, $\bar{\eta}^P_k=\frac{1}{N}\sum_{i}\eta^P_{ik}$, and scalars $\zeta^Q_{ik}$,
$\eta^Q_{ik}$, and $\bar{\eta}^Q_k$ are defined analogously.

While it is possible to define (hard)~accuracy for this multilabel case, a more
common evaluation metric is the micro F\textsubscript{1} score, which in our
notation can be expressed as:
\begin{equation*}
  m(\mathbf{P},\mathbf{Q})\eqdef2\frac{\sum_{ik}J^P_{ik}J^Q_{ik}}{\sum_{ik}\left(J^P_{ik}+J^Q_{ik} \right)}
\end{equation*}
where $J^P_{ik}=\mathds{1}(P_{ik}>0.5)$, and $J^Q_{ik}$ is defined analogously.
Similarly, we define the soft version of the micro F\textsubscript{1} score as
follows~(see derivations in the Appendix):
\begin{equation*}
  m(\mathbf{P},\mathbf{Q})\eqdef2\frac{\sum_{ik}\min(P_{ik},Q_{ik})}{\sum_{ik}(P_{ik}+Q_{ik})}.
\end{equation*}
The (hard)~micro
F\textsubscript{1} score is a special case of this soft counterpart when
$P_{ik}\in\{0,1\}$ and $Q_{ik}\in\{0,1\}$~(i.e. no label variation).
Furthermore, the soft accuracy in \Cref{sec:metrics} is a special case of this
soft micro F\textsubscript{1} score when the unity constraints hold. Therefore,
\emph{the soft micro F\textsubscript{1} score is a general evaluation metric for
  classification tasks}.

\subsection{Relationship between Metrics}\label{sec:rel-overall-metrics}

To analyse the relationship between metrics, we compute Pearson correlation
between each pair of multiclass metrics~(excluding the macro ones) similar to
prior work~\citep{chicco2020}. For a given number classes, we sample $B$
matrices $\mathbf{P}$ and $\mathbf{Q}$ and compute the value of each metric on
these $B$ pairs of $\mathbf{P}$ and $\mathbf{Q}$. The Pearson correlations
between pairs of metrics are then computed based on these $B$ data points. We
set $B=500$ for computational reasons. We report the correlation coefficients in
\Cref{tbl:pairwise-corr}.

\Cref{tbl:pairwise-corr} shows that the metrics are generally weakly correlated
with each other, with the exception of soft accuracy and \pojsd where the two
are consistently strongly correlated~($r>0.9$). This strong correlation with an
established HLV metric gives an empirical assurance that our proposed soft
accuracy is a sensible metric. The table also shows that soft accuracy is only
moderately correlated with (hard)~accuracy in most cases, where this correlation
gets weaker when the number of classes is large~($K=100$), or either
$\mathbf{P}$ or $\mathbf{Q}$ is dense~($\alpha=10$ or $\beta=10$). This finding
suggests that hard and soft accuracy capture different aspects when humans
disagree. Furthermore, the table shows that entropy correlation is poorly
correlated with all metrics, which suggests that it is an outlier HLV metric.

\paragraph{Soft Accuracy and \pojsd}

\begin{table}
  \caption{Pearson correlations between a pair of evaluation metrics
    $m(\mathbf{P},\mathbf{Q})$ for 1K examples and various number of
    classes~($K$) where the rows of $\mathbf{P}$ and $\mathbf{Q}$ are drawn from
    a symmetric Dirichlet with parameters $\alpha$ and $\beta$ respectively. A,
    J, E, and S denote the accuracy, \pojsd, entropy correlation, and soft
    accuracy respectively.}\label{tbl:pairwise-corr}
  \sisetup{table-auto-round, print-zero-integer=false}
  \begin{tabular}{@{}r*{6}{S[table-format=1.2]}@{}}
    \toprule
    $K$ & {A-J}     & {A-E}     & {A-S}     & {J-E}     & {J-S}    & {E-S}     \\
    \midrule
    \multicolumn{7}{c}{$\alpha=\beta=10$}                                             \\
    \addlinespace
    10  & 0.272347  & 0.050255  & 0.289984  & 0.035927  & 0.942263 & 0.100303  \\
    100 & 0.047588  & 0.040964  & 0.051956  & 0.091122  & 0.924129 & 0.087560  \\
    \cmidrule{1-7}
    \multicolumn{7}{c}{$\alpha=\beta=0.1$}                                            \\
    \addlinespace
    10  & 0.627773  & 0.063820  & 0.733743  & 0.126164  & 0.963214 & 0.162816  \\
    100 & 0.194069  & 0.093113  & 0.251355  & 0.014869  & 0.961086 & 0.032225  \\
    \cmidrule{1-7}
    \multicolumn{7}{c}{$\alpha=10,\beta=0.1$}                                         \\
    \addlinespace
    10  & 0.226175  & -0.015882 & 0.206464  & -0.018527 & 0.975144 & -0.002286 \\
    100 & -0.004257 & -0.062204 & -0.026900 & 0.050080  & 0.957816 & 0.061640  \\
    \cmidrule{1-7}
    \multicolumn{7}{c}{$\alpha=0.1,\beta=10$}                                         \\
    \addlinespace
    10  & 0.211874  & 0.016467  & 0.181706  & -0.016632 & 0.975830 & -0.019553 \\
    100 & 0.103478  & -0.021530 & 0.057268  & -0.005990 & 0.967420 & 0.005447  \\
  \bottomrule
  \end{tabular}
\end{table}

Despite their strong positive correlation, soft accuracy and \pojsd have a
notable difference: soft accuracy is upper-bounded by \pojsd.
\begin{theorem}
  Consider the same definitions of $P_{ik}$ and $Q_{ik}$ used in \Cref{sec:eval-metrics}.
  The soft accuracy and the \pojsd between $\mathbf{P}$ and $\mathbf{Q}$
  satisfy the following inequality:
  \begin{equation*}
    \frac{1}{N}\sum_{ik}\min(P_{ik},Q_{ik})\leq1-\frac{1}{N}\sum_{i}\JSD(\mathbf{p}_i,\mathbf{q}_i)
  \end{equation*}
  where $\JSD(\mathbf{p}_i,\mathbf{q}_i)$ denote the Jensen-Shannon
  divergence between the $i$-th rows of $\mathbf{P}$ and $\mathbf{Q}$.
\end{theorem}

\begin{figure}
  \centering
  \includegraphics[width=0.7\linewidth]{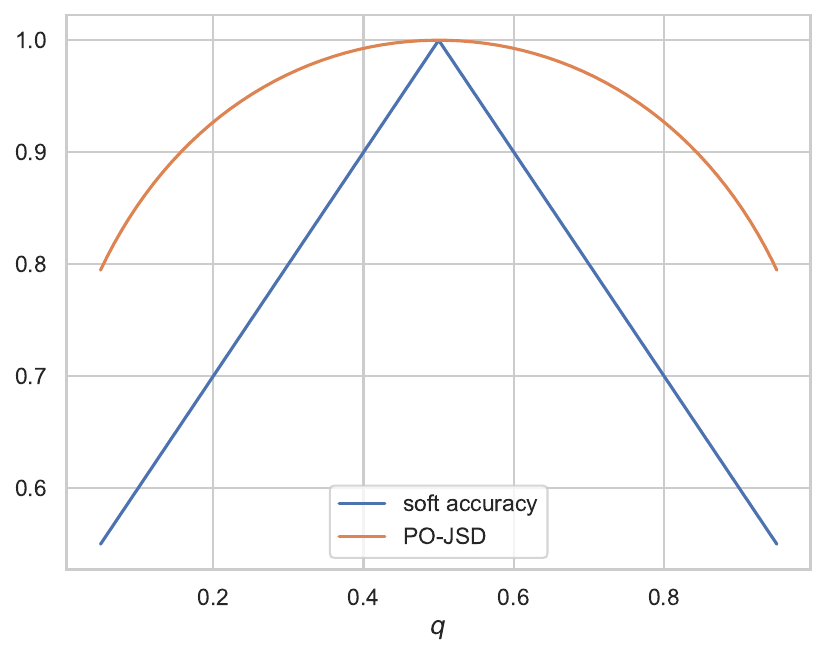}
  \caption{Soft accuracy and \pojsd graphs for a binary classification problem
    on a single example where the true and the predicted judgement for the
    positive class is 0.5 and $q$ respectively.}\label{fig:sa-pojsd}
\end{figure}

\begin{proof}
  Given a scalar $0\leq\pi\leq1$ and discrete distributions
  $\mathbf{a},\mathbf{b}$, it has been shown that the following bound
  holds~\citep[Theorem 4]{lin1991}:
  \begin{equation*}
    \sum_k\min(\pi a_k,(1-\pi)b_k)
    \leq\frac{1}{2}\left(H([\pi,1-\pi])-\JSD_\pi(\mathbf{a},\mathbf{b})\right)
  \end{equation*}
  where $H(\mathbf{u})$ is the entropy of $\mathbf{u}$, and
  $\JSD_\pi(\mathbf{a},\mathbf{b})=H(\pi\mathbf{a}+(1-\pi)\mathbf{b})-\pi
  H(\mathbf{a})-(1-\pi)H(\mathbf{b})$ is the general form of Jensen-Shannon
  divergence with two distributions. It can be easily shown that
  $\JSD_\pi$ is reduced to the standard Jensen-Shannon divergence when
  $\pi=\frac{1}{2}$. Performing this substitution and noting that
  $H([\frac{1}{2},\frac{1}{2}])=1$,\footnote{All logarithms are of base 2.} we
  have
  \begin{equation*}
    \sum_k\min(\frac{1}{2}P_{ik},\frac{1}{2}Q_{ik})\leq\frac{1}{2}\left(1-\JSD(\mathbf{p}_i,\mathbf{q}_i)\right)
  \end{equation*}
  for all $i$. Doubling both sides and taking the average of this inequality
  over $i$ give the desired result.
\end{proof}
This fact implies that soft accuracy penalises a model more heavily than \pojsd,
as illustrated in \Cref{fig:sa-pojsd}. It shows that a skewed predicted
judgement distribution of $(0.2,0.8)$ has a soft accuracy of \num{0.7}, much
less than the \pojsd~($> 0.9$). The high \pojsd is potentially misleading as it
is very close to the maximum possible value of 1.

\section{Training Methods for HLV}

\begin{table}
  \caption{Overview of HLV training methods used, along with details of whether
    annotator identity is required~(I), multilabel tasks are supported~(M), and
    training data is enlarged~(D) which makes training run much
    longer.}\label{tbl:train-meths}
  \begin{tabular}{@{}lccc@{}}
    \toprule
    Name                                                           & I    & M    & D    \\
    \midrule
    Repeated labelling~(\methname{ReL})                            & \no  & \yes & \yes \\
    Majority voting~(\methname{MV})                                & \no  & \yes & \no  \\
    Annotator ranking~(\methname{AR})                              & \yes & \yes & \no  \\
    Annotator ranking~(hard)~(\methname{ARh})                      & \yes & \yes & \no  \\
    Ambiguous labelling~(\methname{AL})                            & \no  & \no  & \yes \\
    Soft labelling~(\methname{SL})                                 & \no  & \yes & \no  \\
    Multitask of \methname{SL} and \methname{MV}~(\methname{SLMV}) & \no  & \yes & \no  \\
    Annotator ensemble~(\methname{AE})                             & \yes & \yes & \no  \\
    Annotator ensemble~(hard)~(\methname{AEh})                     & \yes & \yes & \no  \\
    \cmidrule{1-4}
    \multicolumn{4}{c}{\textit{Below are proposed in this work}}                        \\
    \cmidrule{1-4}
    Jensen-Shannon divergence~(\methname{JSD})                     & \no  & \yes & \no  \\
    Soft micro F\textsubscript{1}~(\methname{SmF1})                & \no  & \yes & \no  \\
    Soft macro F\textsubscript{1}~(\methname{SMF1})                & \no  & \yes & \no  \\
    Loss aggregation with $\min$~(\methname{LA-min})               & \no  & \yes & \no  \\
    Loss aggregation with $\max$~(\methname{LA-max})               & \no  & \yes & \no  \\
    \bottomrule
  \end{tabular}
\end{table}

We now discuss the training methods. In total,
we experiment with 14 HLV training methods, as summarised in
\Cref{tbl:train-meths}. The first method~(\methname{ReL}) is the simplest and
uses standard cross-entropy loss as the training objective based on prior
work~\citep{sheng2008}. The next 8
methods~(\methname{MV}, \methname{AR}, \methname{ARh}, \methname{AL},
\methname{SL}, \methname{SLMV}, \methname{AE}, \methname{AEh}) are from the
Learning with Disagreements shared task~\citep{leonardelli2023} which also use
standard cross-entropy loss. The next 3 methods~(\methname{JSD}, \methname{SmF1},
\methname{SMF1}) use soft metrics that are differentiable as the training
objective. The last 2 methods~(\methname{LA-min}, \methname{LA-max}) aggregate
the cross-entropy losses of all annotations for a given input. In detail:
\begin{enumerate}
  \item \methname{ReL}~(repeated labelling) trains on the disaggregated labels
    directly~\citep{sheng2008}. For example, if the training data only has a
    single instance $x$ with annotations $y_1$ and $y_2$ then \methname{ReL}
    constructs new training data $\{(x,y_1),(x,y_2)\}$.
  \item \methname{MV} trains on the majority-voted labels. This is the baseline
    model given by the shared task organisers.
  \item \methname{AR}~(annotator ranking) and \methname{ARh}~(annotator ranking
    hard) weight each training instance equal to the sum of ranking scores of
    its annotators~\citep{cui2023}. Suppose that training instance $x_1$ is
    labelled as A, A, B, and A by 4 annotators; and $x_2$ is labelled as A, B,
    and B by only the first 3 annotators. Then, to compute the ranking score of
    an annotator:
    \begin{enumerate}
      \item \methname{AR} computes the average judgement of the majority-voted
        label given by the annotator. The annotator ranking scores are~(in
        order) $\frac{3}{4}$, $\frac{1}{2}\left(\frac{3}{4}+\frac{2}{3}
        \right)$, $\frac{2}{3}$, and $\frac{3}{4}$.
      \item \methname{ARh} computes the average number of times the annotator
        agrees with the majority. The scores are $\frac{1}{2}$, $\frac{2}{2}$,
        $\frac{1}{2}$, and $\frac{1}{1}$.
    \end{enumerate}
    For both methods, the ranking score is zero if the annotator never agrees
    with the majority. Given an instance, the training objective is maximising
    the probability of its majority-voted label(s), scaled by its weight as
    defined above.
  \item \methname{AL}~(ambiguous labelling) labels \emph{ambiguous} instances,
    i.e. instances where annotators are not unanimous in their judgement, with
    all selected classes exactly once~\citep{gajewska2023}. Each instance-label
    pair is included in the training set. For example, if training instance
    $x_1$ is labelled as $y_1$, $y_1$, and $y_2$ by 3 annotators, and $x_2$ is
    labelled as $y_2$, $y_2$, and $y_2$, then \methname{AL} produces
    $\lbrace(x_1,y_1),(x_1,y_2),(x_2,y_2)\rbrace$ as the training data.
    Because it was developed for single-label tasks, we exclude this method from
    multilabel tasks.\footnote{Extending this method to multilabel tasks is
      beyond the scope of this work.}
  \item \methname{SL}~(soft labelling) trains on soft labels as
    targets~\citep{maity2023,wan2023}. For example, if an instance is labelled
    as positive by 80\% of annotators then the log-likelihood is $0.8\log
    p(1)+0.2\log p(0)$, where $p(1)$ and $p(0)$ denote the predicted probability
    of the positive and the negative class respectively.
  \item \methname{SLMV} performs multi-task learning using both the
    soft~(\methname{SL}) and the majority-voted~(\methname{MV}) labels as
    targets~\citep{grotzinger2023}. Two models with a shared encoder are trained
    to predict the two types of labels respectively, which is similar to the
    method of \citet{fornaciari2021}. For example, if an instance is labelled as
    positive by 80\% of annotators then the multitask log-likelihood is $0.8\log
    p_{\theta_1}(1)+0.2\log p_{\theta_1}(0)+\log p_{\theta_2}(1)$, where
    $\theta_1$ and $\theta_2$ denote the parameters of the first and the second
    model respectively. We predict only the soft labels for evaluation.
  \item \methname{AE}~(annotator ensembling) and \methname{AEh}~(annotator
    ensembling hard) model each annotator separately followed by
    ensembling~\citep{sullivan2023,vitsakis2023}. In this approach, we have as
    many models as there are annotators. Each model is trained on the labels
    given by a distinct annotator. In practice, we share the encoder parameters
    across models. Suppose that there are 3 annotators, and the
    predicted judgements of the positive class for an instance are 60\%, 30\%,
    and 90\%. To ensemble these predictions at test time:
    \begin{enumerate}
      \item \methname{AE} computes a simple mean~\citep{sullivan2023}. The final
        predicted judgement for the positive class is
        $\frac{1}{3}\left(0.6+0.3+0.9\right)$.
      \item \methname{AEh} computes the distribution of most likely
        labels~\citep{vitsakis2023}. The final predicted judgement for the
        positive class is $\frac{2}{3}$ because 2 out of 3 judgements are over
        50\%.
    \end{enumerate}
  \item \methname{JSD}, \methname{SmF1}~(soft micro F\textsubscript{1}), and
    \methname{SMF1}~(soft macro F\textsubscript{1}) train using JSD, $1-$ soft
    accuracy for multiclass or soft micro F\textsubscript{1} for multilabel
    tasks, and $1-$ soft macro F\textsubscript{1} as the objective
    respectively~(see \Cref{sec:eval-metrics} for definitions).\footnote{We
      exclude entropy correlation because it doesn't have a unique maximiser.}
    These approaches leverage the differentiability of soft metrics, resulting
    in a single objective for both training and inference. By directly
    optimising the evaluation metric at training time, we expect the models to
    exhibit superior performance, especially when evaluated with the
    corresponding metric.
  \item \methname{LA-min}~(loss aggregation $\min$) and \methname{LA-max}~(loss
    aggregation $\max$) aggregate the losses of all annotations for a given
    training instance using the $\min$ and the $\max$ functions respectively.
    For example, given a training instance $x$ with annotations $y_1$ and $y_2$,
    we compute the losses $l_1=\mathcal{L}(x,y_1)$ and $l_2=\mathcal{L}(x,y_2)$
    where $\mathcal{L}$ is the cross-entropy loss function. Then, we update
    model parameters based on the gradient of $g(l_1,l_2)$ where $g$ is either
    the $\min$ or the $\max$ function. Using the $\min$~(resp. $\max$) function
    means selecting the least~(resp.\ most) ``surprising'' annotation for the
    model. Using the mean aggregation is mathematically equivalent to
    \methname{SL} and thus excluded.
\end{enumerate}

\section{Evaluation of HLV Training Methods}

We now conduct an extensive study covering 6 datasets, 14 training methods, and 6
evaluation metrics across 2 pretrained models to understand the performance
landscape of HLV data. \Cref{tbl:datasets} shows an overview of the 6 datasets.
These datasets are selected to represent diverse factors such as language, type
of annotators, number of annotations for each example, number of classes,
presence of annotator identity, and type of classification tasks~(i.e. single-
vs multilabel). By using these diverse datasets, consistent patterns that emerge
across datasets are more likely to hold true in general rather than just in a
specific type of datasets.

\subsection{Experimental Setup}

\begin{table}\footnotesize
  \caption{Datasets used in the evaluation of HLV training methods,
    along with the language~(L), type of annotators~(O) where E and C mean
    Experts and Crowds respectively, number of examples to the nearest
    thousand~($N$), average number of annotations per example~($J$), number of
    classes~($K$), whether annotator identity is present~(I), and whether the
    task is multilabel classification~(M).}~\label{tbl:datasets}
  \sisetup{table-format=3.1}
  \begin{tabular}{@{}llllrSrcc@{}}
    \toprule
    Name                                 & Task            & L  & O & $N$ & $J$   & $K$ & I    & M    \\
    \midrule
    HS-Brexit~\citep{akhtar2021a}        & Hate speech     & EN & E & 1   & 6.0   & 2   & \yes & \no  \\
    MD-Agreement~\citep{leonardelli2021} & Offensiveness   & EN & C & 10  & 5.0   & 2   & \yes & \no  \\
    ArMIS~\citep{almanea2022}            & Misogyny        & AR & E & 1   & 3.0   & 2   & \yes & \no  \\
    ChaosNLI~\citep{nie2020}             & NLI             & EN & C & 3   & 100.0 & 3   & \no  & \no  \\
    MFRC~\citep{trager2022a}             & Moral sentiment & EN & E & 18  & 3.4   & 8   & \yes & \yes \\
    TAG~(private dataset)                & Legal area      & EN & E & 11  & 5.5   & 33  & \yes & \yes \\
    \bottomrule
  \end{tabular}
\end{table}

\paragraph{Public datasets}

We use 5 public datasets with disaggregated annotations that have been used in
previous HLV studies, four of which are in English and one in Arabic. For
HS-Brexit, MD-Agreement, and ArMIS, we use the same train--test splits as
\citet{leonardelli2023}. For both ChaosNLI and MFRC, we use 10-fold
cross-validation with 10\% of the training portion used as a development set.
For ChaosNLI, we include only the SNLI and MNLI subsets~(each has 1.5K examples)
because of their standard NLI setup of one premise and one hypothesis.

\paragraph{``Text Annotation Game'' (TAG) dataset}

We also include a private dataset called TAG which consists of English texts
describing legal problems written by non-expert legal help seekers. Because it
is based on real-world confidential legal requests from help-seekers, we are
unable to distribute the dataset. However, we still include the TAG dataset due
to its importance for the meta-evaluation later in \Cref{sec:meta-eval}. We have
access to this dataset because of our collaboration with Justice
Connect,\footnote{\url{https://justiceconnect.org.au}} an Australian public
benevolent institution\footnote{As defined by the Australian
  government:~\url{https://www.acnc.gov.au/charity/charities/4a24f21a-38af-e811-a95e-000d3ad24c60/profile}}
providing free legal assistance to laypeople facing legal problems. Each text in
the dataset was annotated by an average of 5.5 practising lawyers, who each
selected one or more out of the 33 areas of law\footnote{Including a special
  label ``Not a legal issue''.} that applied to the problem. Thus, it is a
multilabel classification task. Example areas of law include
\textit{Neighbourhood disputes} and \textit{Housing and residential tenancies}.
Average inter-annotator agreement~($\alpha$) over the areas of law is
\num{0.454}, which is modest.\footnote{We compute this agreement on random 10\%
  selection of the data due to the high computational cost.} This is because
lawyers often have different interpretations of the same problem due to their
different legal specialisations and years of experience. However, these
different interpretations are all valid as human label variation because these
registered lawyers are subject matter experts who play a crucial part in
interpreting the law. The dataset has a total of 11K texts collected between
July 2020 and early December 2023. We randomly split the dataset 8:1:1 for the
training, development, and test sets respectively.

\paragraph{Models}

We experiment with both small and large language models.
Because HS-Brexit, MD-Agreement, and ArMIS use Twitter as source data, we
use TwHIN-BERT~\citep{zhang2023a} which was pretrained on Twitter data and
supports both English and Arabic. For other datasets, we use
RoBERTa~\citep{liu2019h}. In addition, we experiment with
8B LLaMA~3~\citep{grattafiori2024a} for the English datasets.\footnote{While LLaMA 3
  was trained on multilingual data, its intended use is English only.} We
replace the output layer of these pretrained models with a linear layer with $K$
output units. We use the base versions of both RoBERTa and TwHIN-BERT (with roughly 100M parameters each).

\paragraph{Training}

We implement all methods using FlairNLP~\citep{akbik2019}. For RoBERTa and
TwHIN-BERT models, we tune the learning rate and the batch size using random
search. Because there are multiple evaluation metrics that aren't necessarily
comparable, we select the best hyperparameters based on the geometric mean of
their values\footnote{We transform entropy correlation to a value between 0 and
  1 before taking the mean.} to avoid biasing towards one metric. We exclude
both hard and soft macro F\textsubscript{1} scores from this mean as most
datasets are balanced.
For LLaMA, we use FlairNLP's default hyperparameters\footnote{Learning rate and
  batch size are \num{5e-5} and 32 respectively.} for computational reasons. We
use LoRA~\citep{hu2022} to finetune LLaMA efficiently. All models are finetuned
for 10 epochs. We truncate inputs longer than 512 tokens in the TAG dataset,
affecting only 4\% of instances.
For ChaosNLI and MFRC, we train the final model used for evaluation of each fold
on the concatenation of the training and the development sets.

\paragraph{Evaluation}

When cross-validation is used, we evaluate the concatenated test predictions.
Otherwise, we evaluate on the test set. We perform the evaluation across 3
training runs.

\subsection{Results}

\begin{figure*}
  \centering
  \includegraphics[width=\textwidth]{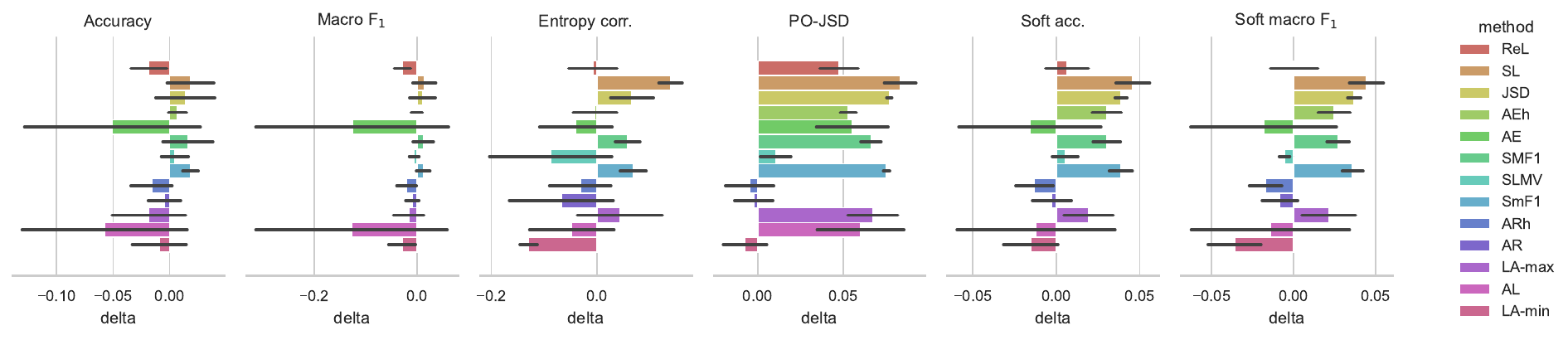}
  \caption{Performance difference~(delta) with mean \methname{MV} performance of
    TwHIN-BERT on ArMIS. The methods are sorted by their mean ranking across datasets,
    models, and metrics.}\label{fig:delta-mv-results-armis}
\end{figure*}
\begin{figure*}
  \centering
  \includegraphics[width=\textwidth]{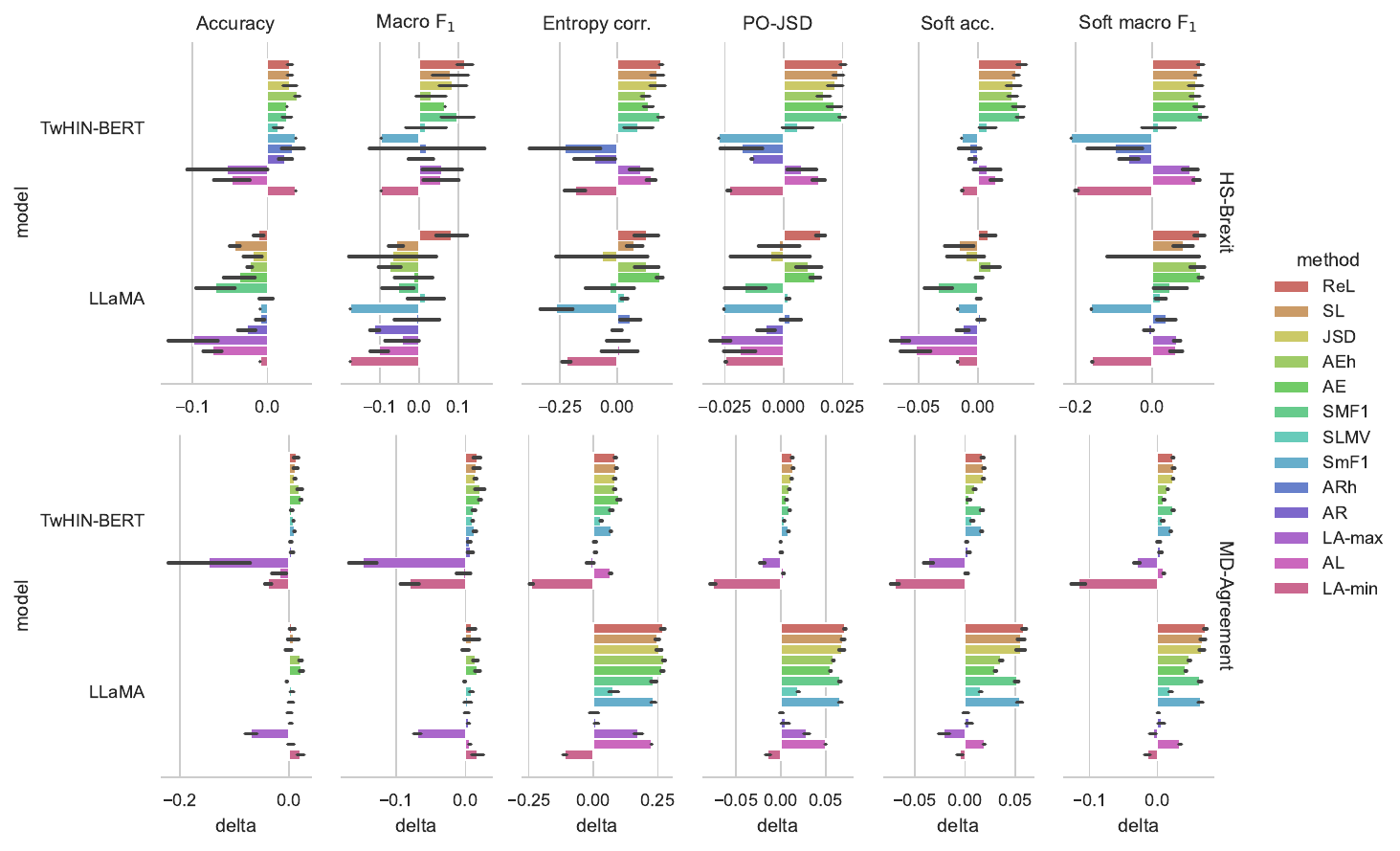}
  \caption{Performance difference~(delta) with mean \methname{MV} performance on
    HS-Brexit and MD-Agreement. \methname{SmF1} with TwHIN-BERT predicts
    zero for all test instances on HS-Brexit so its entropy correlation is
    undefined. The methods are sorted by their mean ranking across datasets,
    models, and metrics.}\label{fig:delta-mv-results-hs-md}
\end{figure*}
\begin{figure*}
  \centering
  \includegraphics[width=\textwidth]{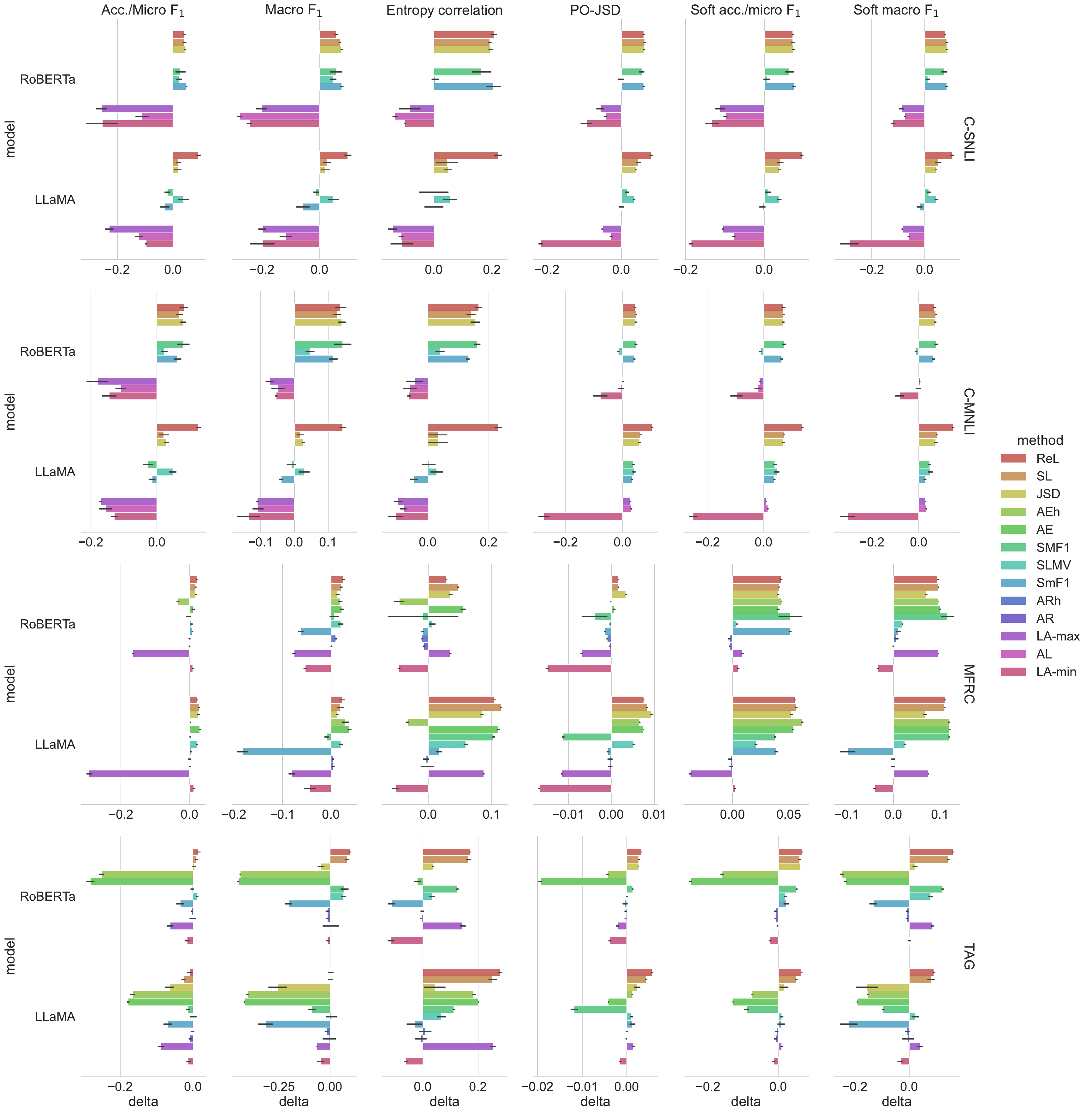}
  \caption{Performance difference~(delta) with mean \methname{MV} performance on
    ChaosNLI~(both the SNLI and the MNLI portions), MFRC, and TAG datasets.
    ChaosNLI doesn't have annotator identity information, so \methname{AE},
    \methname{AEh}, \methname{AR}, \methname{ARh} are inapplicable.
    \methname{AL} is omitted from MFRC and TAG because it is developed for
    single-label rather than multilabel tasks. \methname{AEh} with RoBERTa
    predicts zero for all test instances on several classes in TAG, so its
    entropy correlation is undefined. The methods are sorted by their mean
    ranking across datasets, models, and metrics.}\label{fig:delta-mv-results}
\end{figure*}

We compute the difference between the performance of a single run of an HLV
training method and the mean performance of \methname{MV}. We then report the
mean differences across runs. We present the results on ArMIS in
\Cref{fig:delta-mv-results-armis}~(the only non-English dataset), HS-Brexit and
MD-Agreement in \Cref{fig:delta-mv-results-hs-md}~(Learning with Disagreements
shared task datasets), and the remaining datasets in
\Cref{fig:delta-mv-results}~(multiclass and multilabel datasets). We report
performance of \methname{MV} in \Cref{tbl:mv-results} in the Appendix. We
make the following observations.

First, HLV training methods have different effectiveness compared to
\methname{MV}, with some methods performing very poorly. For example, on ArMIS
and both portions of ChaosNLI, \methname{LA-min} consistently performs worse
than \methname{MV}. This finding suggests the importance of choosing the right
training method when embracing HLV.

Second, both \methname{ReL} and \methname{SL} substantially outperform
\methname{MV} in 68 out of 78 settings~(87.2\% of all
settings),\footnote{Negative cases for \methname{ReL}: all metrics except \pojsd
  on ArMIS~(5 settings), LLaMA on HS-Brexit and MD-Agreement with accuracy~(2),
  LLaMA on MD-Agreement with macro F\textsubscript{1} score~(1), LLaMA on TAG
  with hard metrics~(2). Negative cases for \methname{SL}: TwHIN-BERT on ArMIS
  with hard metrics~(2), LLaMA on HS-Brexit with all metrics except entropy
  correlation and soft macro F\textsubscript{1} score~(4), LLaMA on MD-Agreement
  and TAG with hard metrics~(4).} where a setting is a 3-tuple of a dataset, a
model, and an evaluation metric.\footnote{We consider the SNLI and the MNLI
  portions of ChaosNLI as separate datasets for this purpose.} Furthermore,
\methname{ReL} or \methname{SL} is among the top-performing methods in 89.7\% of
all settings,\footnote{Negative cases: TwHIN-BERT on HS-Brexit with accuracy~(1
  setting); RoBERTa on MFRC with entropy correlation, \pojsd, and soft macro
  F\textsubscript{1} score~(3); LLaMA on MFRC with macro F\textsubscript{1}
  score, \pojsd, and both micro and macro soft F\textsubscript{1} scores~(4).}
including when compared to \methname{JSD} and \methname{SmF1} evaluated using
\pojsd and soft accuracy/micro F\textsubscript{1} respectively. Concretely,
\methname{ReL} and \methname{SL} are on par with~(or outperform in some cases
with LLaMA) the training method whose learning objective is the same as the
evaluation metric in 23 out of 26 settings~(88.5\%),\footnote{We consider only
  \pojsd and soft accuracy/micro F\textsubscript{1} evaluation metrics to get
  the total of 26 settings. Negative cases: RoBERTa on MFRC~(2 settings), LLaMA
  on MFRC with \pojsd~(1).} demonstrating the strength of \methname{ReL} and
\methname{SL}.

Third, \methname{ReL} and \methname{SL} are generally competitive with each
other. However, there are two exceptions to this trend. First, on ArMIS,
\methname{SL} outperforms \methname{ReL}. Second, on both HS-Brexit and
ChaosNLI, \methname{ReL} outperforms \methname{SL} with LLaMA. For the former
finding, however, ArMIS is an outlier dataset: it is the only Arabic dataset,
has the smallest number of examples and annotations per example, and all methods
generally have higher variance on this dataset. The latter finding suggests that
\methname{ReL} is better suited to large language models than \methname{SL}.

Fourth, \methname{SMF1} and \methname{JSD} generally also perform well, although
they lag behind \methname{ReL} and \methname{SL}. Because both
\methname{SMF1} and \methname{JSD} use a soft evaluation metric as the training
objective, this finding underlines the value in using differentiable metrics for
HLV model training.

We also observe that \methname{SmF1} with TwHIN-BERT results in undefined entropy
correlation on HS-Brexit. Looking closely, we find that the method degenerates to
predicting $Q_{ik}=0$ for all $i$. We observe the same degenerate results for
\methname{AEh} with RoBERTa on the TAG dataset. Moreover, \methname{SmF1} with
RoBERTa attains much lower soft micro F\textsubscript{1} score than the
top-performing \methname{ReL} on TAG, even though \methname{SmF1}
optimises the evaluation metric directly during training. Our investigation
suggests that these unexpected observations are due to the class imbalance
present in both HS-Brexit and TAG.\ When the training objective is fair~(e.g.,
\methname{SMF1}), the problem disappears.

\subsection{Discussion}

\methname{ReL} and \methname{SL} appear to perform best as a training method,
and that is somewhat surprising because they are the most straightforward
methods to incorporate HLV in model training.
That said, our finding that \methname{SL} is a strong performer is in line with
what \citet{uma2021} discovered. The results of \methname{ReL}, however, stand
in contrast with prior work~\citep{uma2021,kurniawan2024}.
This difference may be due to the structured prediction tasks that the prior
work considered which have an exponentially large output space. We note that
\citet{uma2021} also found that \methname{ReL} performs really well in simple,
multiclass classification tasks, in line with our findings.

Moreover, the results suggest that \methname{ReL} is likely to outperform
\methname{SL} when large language models are used. A possible explanation is
that with \methname{SL}, the model observes the complete judgement distribution
of an instance as target. Thus, there is not much flexibility as the model has
to predict that distribution to minimise the training loss. In contrast, with
\methname{ReL}, the model only observes one annotation of the instance as
target. Therefore, it has more flexibility in how it distributes the judgement
distribution mass across all the labels of that instance over training steps. We
hypothesise that large language models have sufficient learning capacity to
benefit from this flexibility while smaller models do not. This explanation is
consistent with our findings on ChaosNLI that \methname{ReL} has much larger
gains compared to \methname{SL}. ChaosNLI has 100 annotations per instance, the
largest number out of all datasets. We leave further investigation of this
hypothesis to future work.

Despite its strengths, \methname{ReL} has substantial computational cost:
because it keeps the annotations disaggregated, the size of the training data
can be enormous. For example, in the ChaosNLI dataset, \methname{ReL} makes
the training data 100 times larger because there are exactly 100 annotations for
each training instance. That \methname{ReL} performs better with LLaMA than
RoBERTa on ChaosNLI exacerbates the computational inefficiency of \methname{ReL}:
both the data and the model must be large for it to perform optimally.
Mitigating this inefficiency can be an interesting avenue for future work.

Because the training data is much larger, one may think that the superiority of
\methname{ReL} over \methname{SL} is due to its much longer training. However,
this is not the case. We find that increasing the number of training iterations
of \methname{SL} to match that of \methname{ReL} does not improve performance.

\section{Meta-Evaluation of HLV Evaluation Metrics}\label{sec:meta-eval}

In the previous section, we saw the best training methods that perform
consistently well across metrics. In this section, we address the next question:
what is the best evaluation metric for HLV data? We answer this by conducting a
meta-evaluation on the evaluation metrics. We focus on the TAG dataset
specifically for this meta-evaluation experiment as we have access to
practicing lawyers as annotators who can provide high quality annotations. It
does mean we are unable to distribute the dataset for ethical reasons, but we
believe the insights derived from the meta-evaluation is worth the trade-off.

A good evaluation metric should produce a ranking of training methods
(henceforth ``metric ranking'') that correlates to the ranking produced by human
judgements (henceforth ``human ranking''). To create the latter (human ranking),
we frame it as a pairwise comparison task where we pit two methods against
each other and ask lawyers to select one of them. Given a large number of
human-judged pairwise comparisons, we then use an algorithm to create a ranking
for the training methods. Intuitively, the algorithm will rank a method that
``wins'' consistently higher than another method that ``loses'' most of the time.
Once we have the human ranking, we can then compute its correlation with each
metric ranking, and the best metrics are the ones with the highest correlations.

\subsection{Experimental Setup}

To create the ranking of training methods produced by human judgements, given a
language model (RoBERTa or LLaMA) we randomly sample a pair of methods (e.g.\
\methname{ReL} and \methname{SL}) and present the instance text (i.e.\ a legal
problem) and two results (i.e.\ distribution of the areas of law)\footnote{Note
  that because it is multilabel task, the distribution doesn't sum to one. We
  only show areas of law whose probability exceeds a threshold of \num{0.1}. We
  choose this threshold because it results in similar numbers of areas of law
  whose probability exceeds the value across all methods.} produced by the two
methods, and then ask the lawyers which result is more accurate. The lawyers can
choose either one of them, both, or neither of them. See Appendix
\Cref{fig:pref-annot} for an illustration of the annotation task. On average,
for a given model~(RoBERTa or LLaMA) each pair of methods has 8 judgements made
by 4.4 lawyers across 3--4 randomly-sampled instance texts. This gives a total
of 1.2K pairwise
judgements\footnote{$8\text{ judgements}\times\left(
    \frac{1}{2}\times13\times(13-1) \right)\text{ pairs of methods}\times2\text{
  models}=1248\text{ judgements}$} across 445
instances, 18 lawyers, and 2 models~(RoBERTa and LLaMA). Each pair of instance
text and two results is annotated by an average of 2.3 lawyers.

To make sure the lawyers form their own expectation on the areas of law relevant
to an instance before making this pairwise selection, we first show the instance
text and ask them to select the most relevant areas of law~(\Cref{fig:aol-annot}
in the Appendix). After completing this task for an instance text, we then ask
them to make the pairwise selection. The two tasks must be completed for an
instance text before the lawyers can move to the next instance text. We conduct
several pilot studies with the lawyers to refine the interface design and ensure
the task is clear to them.

At the end of the annotation task, we have a collection of human-judged pairwise
comparisons for the training methods. We next use the rank centrality
algorithm~\citep{negahban2012} to compute the ``score'' for each method to
ultimately produce a ranking of the methods. This algorithm casts the problem as
a random walk on a weighted directed graph where each node corresponds to a
method and the weight of the edge from method $i$ to method $j$ is the
probability that method $j$ wins over model $i$.\footnote{We determine that
  method $j$ wins over method $i$ if the annotator select method $j$ but not
  method $i$.} The score of a method is then its stationary probability for this
random walk. Intuitively, the score is high if it wins against other
high-scoring methods or against many low-scoring methods. To assess the
evaluation metrics and understand which metrics are the best, for each metric we
compute the Pearson correlation between: (1)~method performance as given by the
evaluation metric; and (2)~the method scores produced by human judgements.

\subsection{Results}

\begin{figure}
  \centering
  \includegraphics[width=0.95\linewidth]{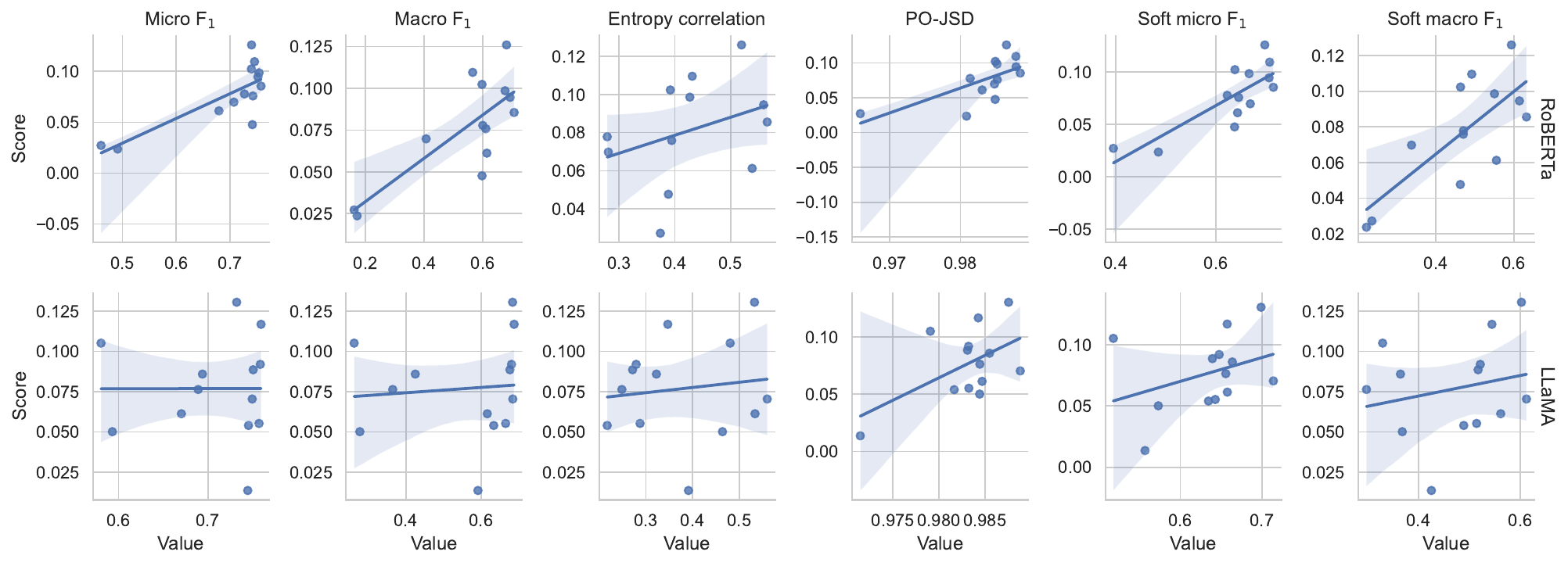}
  \caption{Relationships between method performance as given by the evaluation
    metrics~(Value) and method scores produced by human judgements~(Score) on
    the TAG dataset. The shaded area denotes a 95\% confidence interval. Each
    data point corresponds to a specific training method~(e.g., \methname{ReL}).
    \methname{AEh} is excluded from RoBERTa with entropy correlation because of
    its degenerate results.}\label{fig:rel-human-metric}
\end{figure}

We draw the regression plots between method performance as given by the
evaluation metrics and method scores produced by human judgements in
\Cref{fig:rel-human-metric}. The figure shows that not all metrics exhibit a
strong linear relationship with human judgements. For example, both hard
F\textsubscript{1} scores and entropy correlation show almost no relationship
for LLaMA. All metrics except entropy correlation show a strong linear
relationship for RoBERTa. This suggests that entropy correlation is a poor HLV
metric.

In \Cref{tbl:human-metric-corr}, we show the Pearson correlation coefficients
for all metrics along with the $p$-values. The table confirms that all metrics
are positively correlated with human judgements but with different magnitudes.
As before, both hard F\textsubscript{1} scores and entropy correlation are very
weakly correlated with human judgements for LLaMA, and entropy correlation is
the weakest for RoBERTa. Furthermore, it is the only metric whose correlation is
not statistically significant. Across both pretrained models, both \pojsd and
soft micro F\textsubscript{1} consistently have the strongest correlation, but
each is best for different models: \pojsd is best for LLaMA while soft micro
F\textsubscript{1} score is best for RoBERTa. This finding suggests that both
metrics are good for evaluation in the HLV context.

\subsection{Method ranking based on human judgements}

Given the scores computed from human judgements, we can rank the methods based
on their scores (human ranking).
\Cref{tbl:human-pref-meth-rank} shows that the human rankings have some
similarities to TAG results in \Cref{fig:delta-mv-results}. For instance,
\methname{ReL}, \methname{JSD} and \methname{SL} are generally strong
performers~(exception: \methname{ReL} with LLaMA). That said, we also see some
discrepancies. For example, \methname{SLMV} is highly ranked here but not
reflected in \Cref{fig:delta-mv-results}. We contend, however, that the
meta-evaluation study is ultimately based on one dataset and so we should
interpret these results cautiously.

\begin{table}\small
  \caption{Correlation coefficients between method performance as given by the
    evaluation metrics and method scores produced by human judgements on the TAG
    dataset along with their $p$-values computed with a permutation test.
    Asterisks denote statistical
    significance~($p<.05$).}\label{tbl:human-metric-corr}
  \sisetup{table-format=1.3, print-zero-integer=false}
  \begin{tabular}{@{}lllS[table-space-text-post={$^*$}]S@{}}
    \toprule
    Model                      & Type                  & Metric                        & {Pearson $r$} & {$p$-value} \\
    \midrule
    \multirow{6.2}{*}{RoBERTa} & \multirow{2}{*}{Hard} & Micro F\textsubscript{1}      & 0.787$^*$     & 0.001       \\
                               &                       & Macro F\textsubscript{1}      & 0.774$^*$     & 0.005       \\
    \cmidrule(l){2-5}
                               & \multirow{4}{*}{Soft} & Entropy correlation           & 0.339         & 0.281       \\
                               &                       & \pojsd                        & 0.674$^*$     & 0.005       \\
                               &                       & Soft micro F\textsubscript{1} & 0.804$^*$     & 0.001       \\
                               &                       & Soft macro F\textsubscript{1} & 0.753$^*$     & 0.006       \\
    \cmidrule{1-5}
    \multirow{6.2}{*}{LLaMA}   & \multirow{2}{*}{Hard} & Micro F\textsubscript{1}      & 0.002         & 0.984       \\
                               &                       & Macro F\textsubscript{1}      & 0.087         & 0.776       \\
    \cmidrule(l){2-5}
                               & \multirow{4}{*}{Soft} & Entropy correlation           & 0.125         & 0.689       \\
                               &                       & \pojsd                        & 0.539         & 0.070       \\
                               &                       & Soft micro F\textsubscript{1} & 0.343         & 0.249       \\
                               &                       & Soft macro F\textsubscript{1} & 0.213         & 0.488       \\
    \bottomrule
  \end{tabular}
\end{table}

\begin{table}\small
  \caption{Rankings of HLV training methods based on their scores produced by
    human judgements on the TAG dataset.}\label{tbl:human-pref-meth-rank}
  \sisetup{table-format=1.3, print-zero-integer=false}
  \begin{tabular}{@{}*{4}{lS}@{}}
    \toprule
    \multicolumn{4}{c}{RoBERTa}   & \multicolumn{4}{c}{LLaMA}                                                                       \\
    \cmidrule(r){1-4} \cmidrule(l){5-8}
    \multicolumn{2}{c}{Rank 1--7} & \multicolumn{2}{c}{Rank 8--13} & \multicolumn{2}{c}{Rank 1--7} & \multicolumn{2}{c}{Rank 8--13} \\
    \cmidrule(r){1-2} \cmidrule(lr){3-4} \cmidrule(lr){5-6} \cmidrule(l){7-8}
    Method             & {Score} & Method            & {Score} & Method          & {Score} & Method            & {Score} \\
    \midrule
     \methname{SMF1}   & 0.126   & \methname{MV}     & 0.076   & \methname{SL}   & 0.130   & \methname{ReL}    & 0.070   \\
     \methname{JSD}    & 0.109   & \methname{SmF1}   & 0.070   & \methname{SLMV} & 0.117   & \methname{LA-max} & 0.061   \\
     \methname{ARh}    & 0.102   & \methname{LA-max} & 0.061   & \methname{AE}   & 0.105   & \methname{ARh}    & 0.055   \\
     \methname{SLMV}   & 0.098   & \methname{AR}     & 0.048   & \methname{MV}   & 0.092   & \methname{LA-min} & 0.054   \\
     \methname{SL}     & 0.095   & \methname{AE}     & 0.027   & \methname{AR}   & 0.088   & \methname{AEh}    & 0.050   \\
     \methname{ReL}    & 0.085   & \methname{AEh}    & 0.024   & \methname{JSD}  & 0.086   & \methname{SMF1}   & 0.014   \\
     \methname{LA-min} & 0.078   &                   &         & \methname{SmF1} & 0.076   &                   &         \\
    \bottomrule
  \end{tabular}
\end{table}

\subsection{Discussion}

Putting all the results together, \pojsd and soft micro F\textsubscript{1} are
arguably some of the best metrics for HLV data. That said, \citet{uma2021}
argued that the value of \pojsd is typically confined within a small range and
lacks an intuitive interpretation. We believe that this is its biggest weakness,
which is shared by other information-theoretic measures such as cross-entropy.
In contrast, soft micro F\textsubscript{1} is more interpretable as it is
analogous to the standard F\textsubscript{1} score. Therefore, our general
recommendation is to report both metrics, but focus on the soft micro
F\textsubscript{1} score when an accessible interpretation is important~(e.g.,
communicating with non-technical people) or one is restricted to a single
metric~(e.g., in hyperparameter tuning).

\section{Conclusions}

We propose new evaluation metrics for evaluating model predictions with human
label variation~(HLV). Taking inspiration from remote sensing research, we
represent human judgement distributions as degrees of membership over fuzzy sets
and generalise standard metrics such as accuracy into their soft versions using
fuzzy set operations. Therefore, our proposed metrics have intuitive
interpretations and reduce to standard hard metrics if there is no label
variation. While our analysis shows that our proposed soft accuracy metric is
strongly correlated with an existing metric based on Jensen-Shannon divergence,
we mathematically prove that the former is upper-bounded by the latter. We
further show that the JSD-based metric can produce a misleadingly high score as
a result.

Because the soft metrics are differentiable, we propose 3 new training methods for
HLV using the metrics as the training objective. Additionally, we propose 2 new
training methods that aggregate losses over annotations of the same instance,
bringing the total of our proposed training methods to 5 methods. We
test our proposed methods on 6 datasets spanning binary, multiclass, and
multilabel classification tasks, as well as crowd and expert annotators. We
evaluate against 9 existing HLV training methods across 2 pretrained models and
use a total of 6 HLV evaluation metrics including both existing and proposed
ones to find the best methods for HLV data. Then, we perform an empirical
meta-evaluation of the evaluation metrics to understand which metrics are best
for HLV data.

We find that simple methods such as training on disaggregated
annotations~(\methname{ReL}) or soft labels~(\methname{SL}) perform best in most
cases. They often outperform not only training on the majority-voted labels but
also more complex HLV training methods including our proposed training methods
with the differentiable metrics. Our meta-evaluation shows that our proposed
soft micro F\textsubscript{1} score is one of the best metrics for HLV data.
This metric reduces to our proposed soft accuracy in single-label
classification. Given its intuitive interpretation and positive meta-evaluation
result, we recommend further research to include soft micro F\textsubscript{1}
when reporting model performance in the HLV context.

\section*{Limitations}

The TAG dataset is private and cannot be released publicly, which is a
limitation of our work in terms of reproducibility. This is because the data is
based on real-world confidential legal requests from help-seekers. More
importantly, their safety and privacy is of high concern: some cases are so
unique that even if you were to anonymise the cases, re-identfication may still
be possible. Nevertheless, we believe our work still offers valuable scientific
knowledge on the investigation of HLV training methods and evaluation metrics.

Our empirical meta-evaluation is limited to a single multilabel classification
dataset. Thus, it is unclear if the findings extend to other datasets and binary
or single-label tasks. Future work could expand on this limitation by
experimenting with a different or diverse selection of datasets.

\appendix

\appendixsection{Derivation of Soft F\textsubscript{1} Scores}

\subsection{Fuzzy Sets}

Below we provide a brief overview of fuzzy sets drawn from the work of
\citet{kruse2022}. A fuzzy set $A$ is defined by its membership function
$\mu_A:X\to [0,1]$, where $\mu_A(x)$ represents the degree of membership of
$x\in X$ in $A$. The cardinality of $A$ is defined as the sum of the degrees of
membership of all elements of $X$:
\begin{equation*}
  \sum_{x\in X}\mu_A(x).
\end{equation*}
If $A$ and $B$ are fuzzy sets over the same universe $X$ then their intersection
is a fuzzy set whose membership function is defined as
\begin{equation*}
  \mu_{A\cap B}(x)=t(\mu_A(x),\mu_B(x))
\end{equation*}
where $t$ is a function that satisfies 3 properties: commutativity,
associativity, and monotonicity.\footnote{$\beta\leq\gamma\Rightarrow
  t(\alpha,\beta)\leq t(\alpha,\gamma)$ where $0\leq\alpha,\beta,\gamma\leq1$}
The function $t$ is called a \emph{t-norm~(triangular norm)}. A special t-norm
is the $\min$ function because it is also idempotent, i.e.
$t(\alpha,\alpha)=\alpha$.

\subsection{Soft F\textsubscript{1} Scores}

Let $H_k$ and $R_k$ denote the predicted and the true (crisp)~sets of examples
in class $k$ respectively. Standard precision and recall scores for class $k$
are defined as:
\begin{align*}
  \mathrm{Prec}_k&=\frac{|H_k\cap R_k|}{|H_k|},\\
  \mathrm{Rec}_k&=\frac{|H_k\cap R_k|}{|R_k|}.
\end{align*}

Embracing HLV, we view judgement distributions as degrees of memberships of fuzzy
sets, each corresponds to a class, over a universe of examples. Specifically, we
consider $P_{ik}$ and $Q_{ik}$ as the true and the predicted degrees of
membership of example $i$ in class $k$. Therefore, we can define the soft
precision and recall scores using fuzzy set operations as
follows~\citep{harju2023}:
\begin{equation}
  \mathrm{SoftPrec}_k=\frac{\sum_i\min(P_{ik},Q_{ik})}{\sum_iQ_{ik}}\label{eqn:cw-soft-p}
\end{equation}
and
\begin{equation}
  \mathrm{SoftRec}_k=\frac{\sum_i\min(P_{ik},Q_{ik})}{\sum_iP_{ik}}.\label{eqn:cw-soft-r}
\end{equation}
The soft F\textsubscript{1} score is the harmonic mean between the soft
precision and recall scores as usual:
\begin{equation}
  \mathrm{SoftF\textsubscript{1}}_k=2\frac{\sum_i\min(P_{ik},Q_{ik})}{\sum_i(P_{ik}+Q_{ik})}.\label{eqn:cw-soft-f1}
\end{equation}
Taking the mean of \Cref{eqn:cw-soft-f1} over classes results in the soft macro
F\textsubscript{1} score. To obtain the micro variant, we simply modify the sums
in both the numerator and the denominator of \Cref{eqn:cw-soft-p,eqn:cw-soft-r}
to also iterate over classes to get the soft micro precision and recall scores,
and then take the harmonic mean of the two scores as normal.

\appendixsection{Performance of \methname{MV}}

\Cref{tbl:mv-results} reports the performance of \methname{MV} across datasets
and evaluation metrics.

\begin{table}[!h]\scriptsize
  \caption{Mean~($\pm$ std) performance of \methname{MV} measured by entropy
    correlation, \pojsd, and various versions of F\textsubscript{1} scores.
    MD-Agr refers to MD-Agreement. C-SNLI and C-MNLI refer to the SNLI and the
    MNLI portions of ChaosNLI respectively. Hard~(resp. soft) accuracy
    scores~(for multiclass tasks) are reported in the hard~(resp. soft) micro
    F\textsubscript{1} score column.}\label{tbl:mv-results}
  \begin{tabular}{@{}ll*{6}{c}@{}}
  \toprule
  Dataset                       & Model      & Micro F\textsubscript{1} & Macro F\textsubscript{1} & Entropy corr.   & \pojsd          & Soft micro F\textsubscript{1} & Soft macro F\textsubscript{1} \\
  \midrule
  \multirow[c]{2}{*}{HS-Brexit} & TwHIN-BERT & .903 $\pm$ .009          & .581 $\pm$ .093          & .470 $\pm$ .119 & .941 $\pm$ .010 & .883 $\pm$ .013               & .676 $\pm$ .071               \\
                                & LLaMA      & .950 $\pm$ .012          & .660 $\pm$ .104          & .330 $\pm$ .065 & .939 $\pm$ .002 & .886 $\pm$ .003               & .623 $\pm$ .012               \\
  \cmidrule{1-8}
  \multirow[c]{2}{*}{MD-Agr}    & TwHIN-BERT & .809 $\pm$ .003          & .779 $\pm$ .002          & .379 $\pm$ .002 & .921 $\pm$ .001 & .811 $\pm$ .002               & .790 $\pm$ .003               \\
                                & LLaMA      & .810 $\pm$ .001          & .784 $\pm$ .001          & .215 $\pm$ .019 & .863 $\pm$ .003 & .770 $\pm$ .002               & .743 $\pm$ .003               \\
  \cmidrule{1-8}
  ArMIS                         & TwHIN-BERT & .713 $\pm$ .014          & .705 $\pm$ .014          & .062 $\pm$ .015 & .769 $\pm$ .011 & .702 $\pm$ .009               & .695 $\pm$ .008               \\
  \cmidrule{1-8}
  \multirow[c]{2}{*}{C-SNLI}    & RoBERTa    & .632 $\pm$ .005          & .551 $\pm$ .007          & .123 $\pm$ .014 & .830 $\pm$ .005 & .672 $\pm$ .004               & .629 $\pm$ .007               \\
                                & LLaMA      & .601 $\pm$ .010          & .551 $\pm$ .008          & .118 $\pm$ .022 & .830 $\pm$ .005 & .668 $\pm$ .005               & .633 $\pm$ .007               \\
  \cmidrule{1-8}
  \multirow[c]{2}{*}{C-MNLI}    & RoBERTa    & .496 $\pm$ .014          & .369 $\pm$ .014          & .044 $\pm$ .007 & .854 $\pm$ .009 & .668 $\pm$ .010               & .642 $\pm$ .009               \\
                                & LLaMA      & .514 $\pm$ .017          & .439 $\pm$ .010          & .100 $\pm$ .018 & .827 $\pm$ .005 & .643 $\pm$ .006               & .615 $\pm$ .006               \\
  \cmidrule{1-8}
  \multirow[c]{2}{*}{MFRC}      & RoBERTa    & .657 $\pm$ .001          & .456 $\pm$ .002          & .404 $\pm$ .002 & .954 $\pm$ .000 & .591 $\pm$ .001               & .418 $\pm$ .000               \\
                                & LLaMA      & .649 $\pm$ .001          & .454 $\pm$ .005          & .353 $\pm$ .005 & .948 $\pm$ .000 & .576 $\pm$ .001               & .395 $\pm$ .001               \\
  \cmidrule{1-8}
  \multirow[c]{2}{*}{TAG}       & RoBERTa    & .742 $\pm$ .002          & .610 $\pm$ .014          & .394 $\pm$ .002 & .985 $\pm$ .000 & .645 $\pm$ .002               & .471 $\pm$ .003               \\
                                & LLaMA      & .759 $\pm$ .006          & .681 $\pm$ .005          & .279 $\pm$ .010 & .983 $\pm$ .000 & .648 $\pm$ .002               & .521 $\pm$ .003               \\
  \bottomrule
  \end{tabular}
\end{table}

\appendixsection{Annotation Interface Examples}

\begin{figure}
  \centering
  \includegraphics[width=0.8\textwidth]{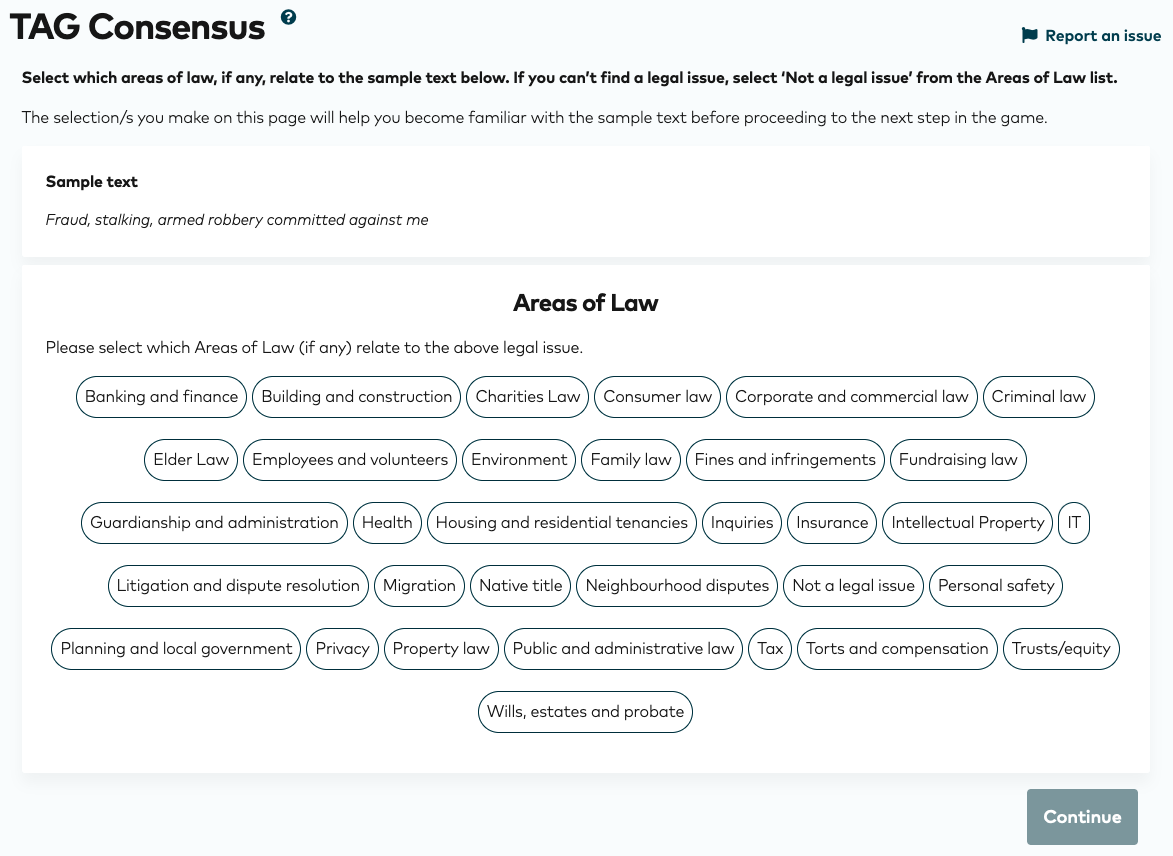}
  \caption{Areas of law annotation interface}\label{fig:aol-annot}
\end{figure}

\begin{figure}
  \centering
  \includegraphics[width=0.8\textwidth]{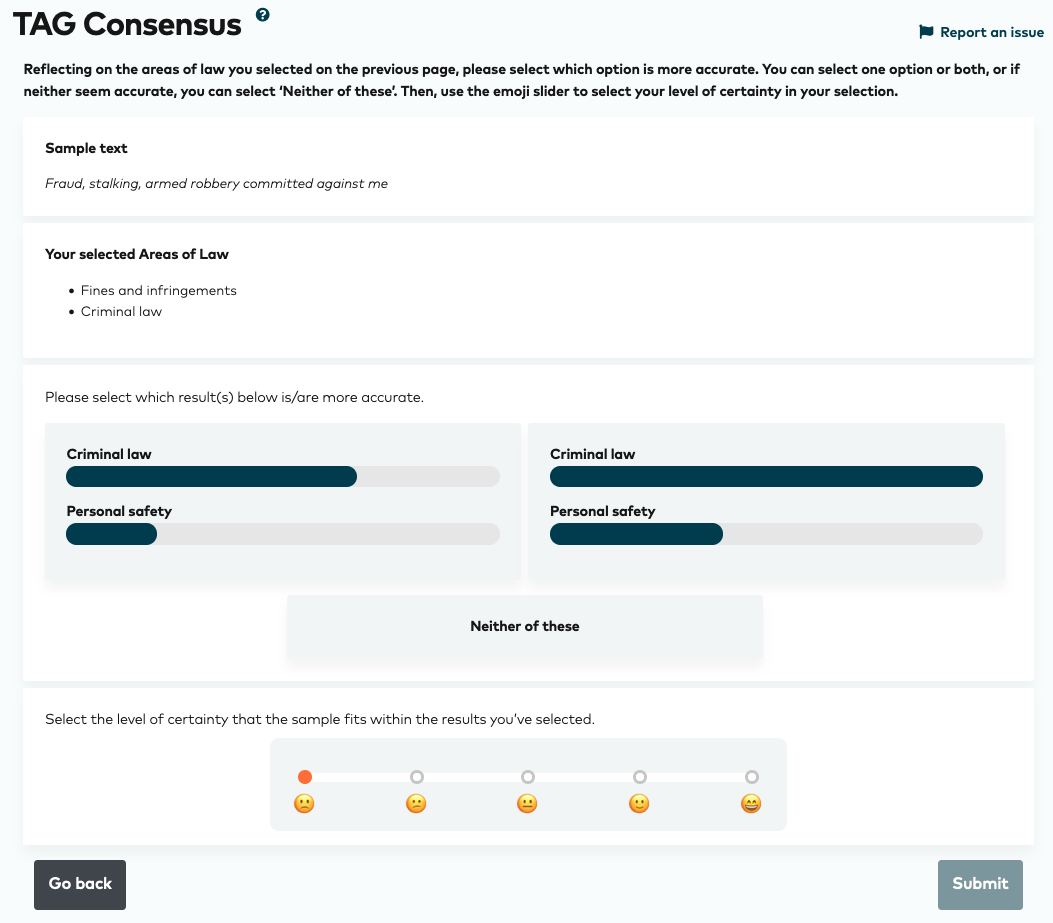}
  \caption{Preference annotation interface}\label{fig:pref-annot}
\end{figure}

\Cref{fig:aol-annot,fig:pref-annot} show the interface of the area of law and
the preference annotation tasks respectively.

\begin{acknowledgments}
This research is supported by the Australian Research Council Linkage Project
LP210200917\footnote{\url{https://dataportal.arc.gov.au/NCGP/Web/Grant/Grant/LP210200917}}
and funded by the Australian Government. This research is done in collaboration
with Justice Connect, an Australian public benevolent institution.\footnote{As
  defined by the Australian
  government:~\url{https://www.acnc.gov.au/charity/charities/4a24f21a-38af-e811-a95e-000d3ad24c60/profile}}
We thank Kate Fazio, Tom O'Doherty, and Rose Hyland from Justice Connect for
their support throughout the project. We thank David McNamara, Yashik Anand, and
Mark Pendergast also from Justice Connect for the development of the
meta-evaluation annotation interface. This research is supported by The
University of Melbourne’s Research Computing Services and the Petascale Campus
Initiative.
\end{acknowledgments}

\bibliographystyle{compling}
\bibliography{em}

\begin{thebibliography}{43}
\expandafter\ifx\csname natexlab\endcsname\relax\def\natexlab#1{#1}\fi

\bibitem[{Akbik et~al.(2019)Akbik, Bergmann, Blythe, Rasul, Schweter, and
  Vollgraf}]{akbik2019}
Akbik, Alan, Tanja Bergmann, Duncan Blythe, Kashif Rasul, Stefan Schweter, and
  Roland Vollgraf. 2019.
\newblock {{FLAIR}}: {{An}} easy-to-use framework for state-of-the-art {{NLP}}.
\newblock In \emph{Proceedings of the 2019 {{Conference}} of the {{North
  American Chapter}} of the {{Association}} for {{Computational Linguistics}}
  ({{Demonstrations}})}, pages 54--59.

\bibitem[{Akhtar, Basile, and Patti(2021)}]{akhtar2021a}
Akhtar, Sohail, Valerio Basile, and Viviana Patti. 2021.
\newblock Whose {{Opinions Matter}}? {{Perspective-aware Models}} to {{Identify
  Opinions}} of {{Hate Speech Victims}} in {{Abusive Language Detection}}.
\newblock \emph{CoRR}, cs.CL/2106.15896.

\bibitem[{Almanea and Poesio(2022)}]{almanea2022}
Almanea, Dina and Massimo Poesio. 2022.
\newblock {{ArMIS}} - {{The Arabic Misogyny}} and {{Sexism Corpus}} with
  {{Annotator Subjective Disagreements}}.
\newblock In \emph{Proceedings of the {{Thirteenth Language Resources}} and
  {{Evaluation Conference}}}, pages 2282--2291.

\bibitem[{Binaghi et~al.(1999)Binaghi, Brivio, Ghezzi, and
  Rampini}]{binaghi1999}
Binaghi, Elisabetta, Pietro~A. Brivio, Paolo Ghezzi, and Anna Rampini. 1999.
\newblock A fuzzy set-based accuracy assessment of soft classification.
\newblock \emph{Pattern Recognition Letters}, 20(9):935--948.

\bibitem[{Chen et~al.(2024)Chen, Wang, Peng, Litschko, Korhonen, and
  Plank}]{chen2024}
Chen, Beiduo, Xinpeng Wang, Siyao Peng, Robert Litschko, Anna Korhonen, and
  Barbara Plank. 2024.
\newblock ``{{Seeing}} the {{Big}} through the {{Small}}'': {{Can LLMs
  Approximate Human Judgment Distributions}} on {{NLI}} from a {{Few
  Explanations}}?
\newblock In \emph{Findings of the {{Association}} for {{Computational
  Linguistics}}: {{EMNLP}} 2024}, pages 14396--14419.

\bibitem[{Chicco and Jurman(2020)}]{chicco2020}
Chicco, Davide and Giuseppe Jurman. 2020.
\newblock The advantages of the {{Matthews}} correlation coefficient ({{MCC}})
  over {{F1}} score and accuracy in binary classification evaluation.
\newblock \emph{BMC Genomics}, 21(1):6.

\bibitem[{Cui(2023)}]{cui2023}
Cui, Xia. 2023.
\newblock Xiacui at {{SemEval-2023 Task}} 11: {{Learning}} a {{Model}} in
  {{Mixed-Annotator Datasets Using Annotator Ranking Scores}} as {{Training
  Weights}}.
\newblock In \emph{Proceedings of the 17th {{International Workshop}} on
  {{Semantic Evaluation}} ({{SemEval-2023}})}, pages 1076--1084.

\bibitem[{Deng et~al.(2023)Deng, Zhang, Liu, Wu, Wang, and Mihalcea}]{deng2023}
Deng, Naihao, Xinliang Zhang, Siyang Liu, Winston Wu, Lu~Wang, and Rada
  Mihalcea. 2023.
\newblock You {{Are What You Annotate}}: {{Towards Better Models}} through
  {{Annotator Representations}}.
\newblock In \emph{Findings of the {{Association}} for {{Computational
  Linguistics}}: {{EMNLP}} 2023}, pages 12475--12498.

\bibitem[{Foody(1996)}]{foody1996}
Foody, G.~M. 1996.
\newblock Approaches for the production and evaluation of fuzzy land cover
  classifications from remotely-sensed data.
\newblock \emph{International Journal of Remote Sensing}, 17(7):1317--1340.

\bibitem[{Fornaciari et~al.(2021)Fornaciari, Uma, Paun, Plank, Hovy, and
  Poesio}]{fornaciari2021}
Fornaciari, Tommaso, Alexandra Uma, Silviu Paun, Barbara Plank, Dirk Hovy, and
  Massimo Poesio. 2021.
\newblock Beyond {{Black}} \& {{White}}: {{Leveraging Annotator Disagreement}}
  via {{Soft-Label Multi-Task Learning}}.
\newblock In \emph{Proceedings of the 2021 {{Conference}} of the {{North
  American Chapter}} of the {{Association}} for {{Computational Linguistics}}:
  {{Human Language Technologies}}}, pages 2591--2597.

\bibitem[{Gajewska(2023)}]{gajewska2023}
Gajewska, Ewelina. 2023.
\newblock Eevvgg at {{SemEval-2023 Task}} 11: {{Offensive Language
  Classification}} with {{Rater-based Information}}.
\newblock In \emph{Proceedings of the 17th {{International Workshop}} on
  {{Semantic Evaluation}} ({{SemEval-2023}})}, pages 171--176.

\bibitem[{G{\'o}mez, Biging, and Montero(2008)}]{gomez2008}
G{\'o}mez, D., G.~Biging, and J.~Montero. 2008.
\newblock Accuracy statistics for judging soft classification.
\newblock \emph{International Journal of Remote Sensing}, 29(3):693--709.

\bibitem[{Grattafiori et~al.(2024)Grattafiori, Dubey, Jauhri, Pandey, Kadian,
  {Al-Dahle}, Letman, Mathur, Schelten, Vaughan, Yang, Fan, Goyal, Hartshorn,
  Yang, Mitra, Sravankumar, Korenev, Hinsvark, Rao, Zhang, Rodriguez,
  Gregerson, Spataru, Roziere, Biron, Tang, Chern, Caucheteux, Nayak, Bi,
  Marra, McConnell, Keller, Touret, Wu, Wong, Ferrer, Nikolaidis, Allonsius,
  Song, Pintz, Livshits, Wyatt, Esiobu, Choudhary, Mahajan, {Garcia-Olano},
  Perino, Hupkes, Lakomkin, AlBadawy, Lobanova, Dinan, Smith, Radenovic,
  Guzm{\'a}n, Zhang, Synnaeve, Lee, Anderson, Thattai, Nail, Mialon, Pang,
  Cucurell, Nguyen, Korevaar, Xu, Touvron, Zarov, Ibarra, Kloumann, Misra,
  Evtimov, Zhang, Copet, Lee, Geffert, Vranes, Park, Mahadeokar, Shah, van~der
  Linde, Billock, Hong, Lee, Fu, Chi, Huang, Liu, Wang, Yu, Bitton, Spisak,
  Park, Rocca, Johnstun, Saxe, Jia, Alwala, Prasad, Upasani, Plawiak, Li,
  Heafield, Stone, {El-Arini}, Iyer, Malik, Chiu, Bhalla, Lakhotia,
  {Rantala-Yeary}, van~der Maaten, Chen, Tan, Jenkins, Martin, Madaan, Malo,
  Blecher, Landzaat, de~Oliveira, Muzzi, Pasupuleti, Singh, Paluri, Kardas,
  Tsimpoukelli, Oldham, Rita, Pavlova, Kambadur, Lewis, Si, Singh, Hassan,
  Goyal, Torabi, Bashlykov, Bogoychev, Chatterji, Zhang, Duchenne, {\c C}elebi,
  Alrassy, Zhang, Li, Vasic, Weng, Bhargava, Dubal, Krishnan, Koura, Xu, He,
  Dong, Srinivasan, Ganapathy, Calderer, Cabral, Stojnic, Raileanu, Maheswari,
  Girdhar, Patel, Sauvestre, Polidoro, Sumbaly, Taylor, Silva, Hou, Wang,
  Hosseini, Chennabasappa, Singh, Bell, Kim, Edunov, Nie, Narang, Raparthy,
  Shen, Wan, Bhosale, Zhang, Vandenhende, Batra, Whitman, Sootla, Collot,
  Gururangan, Borodinsky, Herman, Fowler, Sheasha, Georgiou, Scialom,
  Speckbacher, Mihaylov, Xiao, Karn, Goswami, Gupta, Ramanathan, Kerkez,
  Gonguet, Do, Vogeti, Albiero, Petrovic, Chu, Xiong, Fu, Meers, Martinet,
  Wang, Wang, Tan, Xia, Xie, Jia, Wang, Goldschlag, Gaur, Babaei, Wen, Song,
  Zhang, Li, Mao, Coudert, Yan, Chen, Papakipos, Singh, Srivastava, Jain,
  Kelsey, Shajnfeld, Gangidi, Victoria, Goldstand, Menon, Sharma, Boesenberg,
  Baevski, Feinstein, Kallet, Sangani, Teo, Yunus, Lupu, Alvarado, Caples, Gu,
  Ho, Poulton, Ryan, Ramchandani, Dong, Franco, Goyal, Saraf, Chowdhury,
  Gabriel, Bharambe, Eisenman, Yazdan, James, Maurer, Leonhardi, Huang, Loyd,
  Paola, Paranjape, Liu, Wu, Ni, Hancock, Wasti, Spence, Stojkovic, Gamido,
  Montalvo, Parker, Burton, Mejia, Liu, Wang, Kim, Zhou, Hu, Chu, Cai, Tindal,
  Feichtenhofer, Gao, Civin, Beaty, Kreymer, Li, Adkins, Xu, Testuggine, David,
  Parikh, Liskovich, Foss, Wang, Le, Holland, Dowling, Jamil, Montgomery,
  Presani, Hahn, Wood, Le, Brinkman, Arcaute, Dunbar, Smothers, Sun, Kreuk,
  Tian, Kokkinos, Ozgenel, Caggioni, Kanayet, Seide, Florez, Schwarz, Badeer,
  Swee, Halpern, Herman, Sizov, Guangyi, Zhang, Lakshminarayanan, Inan,
  Shojanazeri, Zou, Wang, Zha, Habeeb, Rudolph, Suk, Aspegren, Goldman, Zhan,
  Damlaj, Molybog, Tufanov, Leontiadis, Veliche, Gat, Weissman, Geboski, Kohli,
  Lam, Asher, Gaya, Marcus, Tang, Chan, Zhen, Reizenstein, Teboul, Zhong, Jin,
  Yang, Cummings, Carvill, Shepard, McPhie, Torres, Ginsburg, Wang, Wu, U,
  Saxena, Khandelwal, Zand, Matosich, Veeraraghavan, Michelena, Li, Jagadeesh,
  Huang, Chawla, Huang, Chen, Garg, A, Silva, Bell, Zhang, Guo, Yu, Moshkovich,
  Wehrstedt, Khabsa, Avalani, Bhatt, Mankus, Hasson, Lennie, Reso, Groshev,
  Naumov, Lathi, Keneally, Liu, Seltzer, Valko, Restrepo, Patel, Vyatskov,
  Samvelyan, Clark, Macey, Wang, Hermoso, Metanat, Rastegari, Bansal,
  Santhanam, Parks, White, Bawa, Singhal, Egebo, Usunier, Mehta, Laptev, Dong,
  Cheng, Chernoguz, Hart, Salpekar, Kalinli, Kent, Parekh, Saab, Balaji,
  Rittner, Bontrager, Roux, Dollar, Zvyagina, Ratanchandani, Yuvraj, Liang,
  Alao, Rodriguez, Ayub, Murthy, Nayani, Mitra, Parthasarathy, Li, Hogan,
  Battey, Wang, Howes, Rinott, Mehta, Siby, Bondu, Datta, Chugh, Hunt, Dhillon,
  Sidorov, Pan, Mahajan, Verma, Yamamoto, Ramaswamy, Lindsay, Lindsay, Feng,
  Lin, Zha, Patil, Shankar, Zhang, Zhang, Wang, Agarwal, Sajuyigbe, Chintala,
  Max, Chen, Kehoe, Satterfield, Govindaprasad, Gupta, Deng, Cho, Virk,
  Subramanian, Choudhury, Goldman, Remez, Glaser, Best, Koehler, Robinson, Li,
  Zhang, Matthews, Chou, Shaked, Vontimitta, Ajayi, Montanez, Mohan, Kumar,
  Mangla, Ionescu, Poenaru, Mihailescu, Ivanov, Li, Wang, Jiang, Bouaziz,
  Constable, Tang, Wu, Wang, Wu, Gao, Kleinman, Chen, Hu, Jia, Qi, Li, Zhang,
  Zhang, Adi, Nam, Yu, Wang, Zhao, Hao, Qian, Li, He, Rait, DeVito, Rosnbrick,
  Wen, Yang, Zhao, and Ma}]{grattafiori2024a}
Grattafiori, Aaron, Abhimanyu Dubey, Abhinav Jauhri, Abhinav Pandey, Abhishek
  Kadian, Ahmad {Al-Dahle}, Aiesha Letman, Akhil Mathur, Alan Schelten, Alex
  Vaughan, Amy Yang, Angela Fan, Anirudh Goyal, Anthony Hartshorn, Aobo Yang,
  Archi Mitra, Archie Sravankumar, Artem Korenev, Arthur Hinsvark, Arun Rao,
  Aston Zhang, Aurelien Rodriguez, Austen Gregerson, Ava Spataru, Baptiste
  Roziere, Bethany Biron, Binh Tang, Bobbie Chern, Charlotte Caucheteux, Chaya
  Nayak, Chloe Bi, Chris Marra, Chris McConnell, Christian Keller, Christophe
  Touret, Chunyang Wu, Corinne Wong, Cristian~Canton Ferrer, Cyrus Nikolaidis,
  Damien Allonsius, Daniel Song, Danielle Pintz, Danny Livshits, Danny Wyatt,
  David Esiobu, Dhruv Choudhary, Dhruv Mahajan, Diego {Garcia-Olano}, Diego
  Perino, Dieuwke Hupkes, Egor Lakomkin, Ehab AlBadawy, Elina Lobanova, Emily
  Dinan, Eric~Michael Smith, Filip Radenovic, Francisco Guzm{\'a}n, Frank
  Zhang, Gabriel Synnaeve, Gabrielle Lee, Georgia~Lewis Anderson, Govind
  Thattai, Graeme Nail, Gregoire Mialon, Guan Pang, Guillem Cucurell, Hailey
  Nguyen, Hannah Korevaar, Hu~Xu, Hugo Touvron, Iliyan Zarov, Imanol~Arrieta
  Ibarra, Isabel Kloumann, Ishan Misra, Ivan Evtimov, Jack Zhang, Jade Copet,
  Jaewon Lee, Jan Geffert, Jana Vranes, Jason Park, Jay Mahadeokar, Jeet Shah,
  Jelmer van~der Linde, Jennifer Billock, Jenny Hong, Jenya Lee, Jeremy Fu,
  Jianfeng Chi, Jianyu Huang, Jiawen Liu, Jie Wang, Jiecao Yu, Joanna Bitton,
  Joe Spisak, Jongsoo Park, Joseph Rocca, Joshua Johnstun, Joshua Saxe, Junteng
  Jia, Kalyan~Vasuden Alwala, Karthik Prasad, Kartikeya Upasani, Kate Plawiak,
  Ke~Li, Kenneth Heafield, Kevin Stone, Khalid {El-Arini}, Krithika Iyer,
  Kshitiz Malik, Kuenley Chiu, Kunal Bhalla, Kushal Lakhotia, Lauren
  {Rantala-Yeary}, Laurens van~der Maaten, Lawrence Chen, Liang Tan, Liz
  Jenkins, Louis Martin, Lovish Madaan, Lubo Malo, Lukas Blecher, Lukas
  Landzaat, Luke de~Oliveira, Madeline Muzzi, Mahesh Pasupuleti, Mannat Singh,
  Manohar Paluri, Marcin Kardas, Maria Tsimpoukelli, Mathew Oldham, Mathieu
  Rita, Maya Pavlova, Melanie Kambadur, Mike Lewis, Min Si, Mitesh~Kumar Singh,
  Mona Hassan, Naman Goyal, Narjes Torabi, Nikolay Bashlykov, Nikolay
  Bogoychev, Niladri Chatterji, Ning Zhang, Olivier Duchenne, Onur {\c C}elebi,
  Patrick Alrassy, Pengchuan Zhang, Pengwei Li, Petar Vasic, Peter Weng,
  Prajjwal Bhargava, Pratik Dubal, Praveen Krishnan, Punit~Singh Koura, Puxin
  Xu, Qing He, Qingxiao Dong, Ragavan Srinivasan, Raj Ganapathy, Ramon
  Calderer, Ricardo~Silveira Cabral, Robert Stojnic, Roberta Raileanu, Rohan
  Maheswari, Rohit Girdhar, Rohit Patel, Romain Sauvestre, Ronnie Polidoro,
  Roshan Sumbaly, Ross Taylor, Ruan Silva, Rui Hou, Rui Wang, Saghar Hosseini,
  Sahana Chennabasappa, Sanjay Singh, Sean Bell, Seohyun~Sonia Kim, Sergey
  Edunov, Shaoliang Nie, Sharan Narang, Sharath Raparthy, Sheng Shen, Shengye
  Wan, Shruti Bhosale, Shun Zhang, Simon Vandenhende, Soumya Batra, Spencer
  Whitman, Sten Sootla, Stephane Collot, Suchin Gururangan, Sydney Borodinsky,
  Tamar Herman, Tara Fowler, Tarek Sheasha, Thomas Georgiou, Thomas Scialom,
  Tobias Speckbacher, Todor Mihaylov, Tong Xiao, Ujjwal Karn, Vedanuj Goswami,
  Vibhor Gupta, Vignesh Ramanathan, Viktor Kerkez, Vincent Gonguet, Virginie
  Do, Vish Vogeti, V{\'i}tor Albiero, Vladan Petrovic, Weiwei Chu, Wenhan
  Xiong, Wenyin Fu, Whitney Meers, Xavier Martinet, Xiaodong Wang, Xiaofang
  Wang, Xiaoqing~Ellen Tan, Xide Xia, Xinfeng Xie, Xuchao Jia, Xuewei Wang,
  Yaelle Goldschlag, Yashesh Gaur, Yasmine Babaei, Yi~Wen, Yiwen Song, Yuchen
  Zhang, Yue Li, Yuning Mao, Zacharie~Delpierre Coudert, Zheng Yan, Zhengxing
  Chen, Zoe Papakipos, Aaditya Singh, Aayushi Srivastava, Abha Jain, Adam
  Kelsey, Adam Shajnfeld, Adithya Gangidi, Adolfo Victoria, Ahuva Goldstand,
  Ajay Menon, Ajay Sharma, Alex Boesenberg, Alexei Baevski, Allie Feinstein,
  Amanda Kallet, Amit Sangani, Amos Teo, Anam Yunus, Andrei Lupu, Andres
  Alvarado, Andrew Caples, Andrew Gu, Andrew Ho, Andrew Poulton, Andrew Ryan,
  Ankit Ramchandani, Annie Dong, Annie Franco, Anuj Goyal, Aparajita Saraf,
  Arkabandhu Chowdhury, Ashley Gabriel, Ashwin Bharambe, Assaf Eisenman, Azadeh
  Yazdan, Beau James, Ben Maurer, Benjamin Leonhardi, Bernie Huang, Beth Loyd,
  Beto~De Paola, Bhargavi Paranjape, Bing Liu, Bo~Wu, Boyu Ni, Braden Hancock,
  Bram Wasti, Brandon Spence, Brani Stojkovic, Brian Gamido, Britt Montalvo,
  Carl Parker, Carly Burton, Catalina Mejia, Ce~Liu, Changhan Wang, Changkyu
  Kim, Chao Zhou, Chester Hu, Ching-Hsiang Chu, Chris Cai, Chris Tindal,
  Christoph Feichtenhofer, Cynthia Gao, Damon Civin, Dana Beaty, Daniel
  Kreymer, Daniel Li, David Adkins, David Xu, Davide Testuggine, Delia David,
  Devi Parikh, Diana Liskovich, Didem Foss, Dingkang Wang, Duc Le, Dustin
  Holland, Edward Dowling, Eissa Jamil, Elaine Montgomery, Eleonora Presani,
  Emily Hahn, Emily Wood, Eric-Tuan Le, Erik Brinkman, Esteban Arcaute, Evan
  Dunbar, Evan Smothers, Fei Sun, Felix Kreuk, Feng Tian, Filippos Kokkinos,
  Firat Ozgenel, Francesco Caggioni, Frank Kanayet, Frank Seide,
  Gabriela~Medina Florez, Gabriella Schwarz, Gada Badeer, Georgia Swee, Gil
  Halpern, Grant Herman, Grigory Sizov, Guangyi, Zhang, Guna Lakshminarayanan,
  Hakan Inan, Hamid Shojanazeri, Han Zou, Hannah Wang, Hanwen Zha, Haroun
  Habeeb, Harrison Rudolph, Helen Suk, Henry Aspegren, Hunter Goldman, Hongyuan
  Zhan, Ibrahim Damlaj, Igor Molybog, Igor Tufanov, Ilias Leontiadis,
  Irina-Elena Veliche, Itai Gat, Jake Weissman, James Geboski, James Kohli,
  Janice Lam, Japhet Asher, Jean-Baptiste Gaya, Jeff Marcus, Jeff Tang,
  Jennifer Chan, Jenny Zhen, Jeremy Reizenstein, Jeremy Teboul, Jessica Zhong,
  Jian Jin, Jingyi Yang, Joe Cummings, Jon Carvill, Jon Shepard, Jonathan
  McPhie, Jonathan Torres, Josh Ginsburg, Junjie Wang, Kai Wu, Kam~Hou U, Karan
  Saxena, Kartikay Khandelwal, Katayoun Zand, Kathy Matosich, Kaushik
  Veeraraghavan, Kelly Michelena, Keqian Li, Kiran Jagadeesh, Kun Huang, Kunal
  Chawla, Kyle Huang, Lailin Chen, Lakshya Garg, Lavender A, Leandro Silva, Lee
  Bell, Lei Zhang, Liangpeng Guo, Licheng Yu, Liron Moshkovich, Luca Wehrstedt,
  Madian Khabsa, Manav Avalani, Manish Bhatt, Martynas Mankus, Matan Hasson,
  Matthew Lennie, Matthias Reso, Maxim Groshev, Maxim Naumov, Maya Lathi,
  Meghan Keneally, Miao Liu, Michael~L. Seltzer, Michal Valko, Michelle
  Restrepo, Mihir Patel, Mik Vyatskov, Mikayel Samvelyan, Mike Clark, Mike
  Macey, Mike Wang, Miquel~Jubert Hermoso, Mo~Metanat, Mohammad Rastegari,
  Munish Bansal, Nandhini Santhanam, Natascha Parks, Natasha White, Navyata
  Bawa, Nayan Singhal, Nick Egebo, Nicolas Usunier, Nikhil Mehta,
  Nikolay~Pavlovich Laptev, Ning Dong, Norman Cheng, Oleg Chernoguz, Olivia
  Hart, Omkar Salpekar, Ozlem Kalinli, Parkin Kent, Parth Parekh, Paul Saab,
  Pavan Balaji, Pedro Rittner, Philip Bontrager, Pierre Roux, Piotr Dollar,
  Polina Zvyagina, Prashant Ratanchandani, Pritish Yuvraj, Qian Liang, Rachad
  Alao, Rachel Rodriguez, Rafi Ayub, Raghotham Murthy, Raghu Nayani, Rahul
  Mitra, Rangaprabhu Parthasarathy, Raymond Li, Rebekkah Hogan, Robin Battey,
  Rocky Wang, Russ Howes, Ruty Rinott, Sachin Mehta, Sachin Siby, Sai~Jayesh
  Bondu, Samyak Datta, Sara Chugh, Sara Hunt, Sargun Dhillon, Sasha Sidorov,
  Satadru Pan, Saurabh Mahajan, Saurabh Verma, Seiji Yamamoto, Sharadh
  Ramaswamy, Shaun Lindsay, Shaun Lindsay, Sheng Feng, Shenghao Lin,
  Shengxin~Cindy Zha, Shishir Patil, Shiva Shankar, Shuqiang Zhang, Shuqiang
  Zhang, Sinong Wang, Sneha Agarwal, Soji Sajuyigbe, Soumith Chintala,
  Stephanie Max, Stephen Chen, Steve Kehoe, Steve Satterfield, Sudarshan
  Govindaprasad, Sumit Gupta, Summer Deng, Sungmin Cho, Sunny Virk, Suraj
  Subramanian, Sy~Choudhury, Sydney Goldman, Tal Remez, Tamar Glaser, Tamara
  Best, Thilo Koehler, Thomas Robinson, Tianhe Li, Tianjun Zhang, Tim Matthews,
  Timothy Chou, Tzook Shaked, Varun Vontimitta, Victoria Ajayi, Victoria
  Montanez, Vijai Mohan, Vinay~Satish Kumar, Vishal Mangla, Vlad Ionescu, Vlad
  Poenaru, Vlad~Tiberiu Mihailescu, Vladimir Ivanov, Wei Li, Wenchen Wang,
  Wenwen Jiang, Wes Bouaziz, Will Constable, Xiaocheng Tang, Xiaojian Wu,
  Xiaolan Wang, Xilun Wu, Xinbo Gao, Yaniv Kleinman, Yanjun Chen, Ye~Hu,
  Ye~Jia, Ye~Qi, Yenda Li, Yilin Zhang, Ying Zhang, Yossi Adi, Youngjin Nam,
  Yu, Wang, Yu~Zhao, Yuchen Hao, Yundi Qian, Yunlu Li, Yuzi He, Zach Rait,
  Zachary DeVito, Zef Rosnbrick, Zhaoduo Wen, Zhenyu Yang, Zhiwei Zhao, and
  Zhiyu Ma. 2024.
\newblock The {{Llama}} 3 {{Herd}} of {{Models}}.
\newblock \emph{CoRR}, cs.AI/2407.21783.

\bibitem[{Gr{\"o}tzinger, Heuschkel, and Drews(2023)}]{grotzinger2023}
Gr{\"o}tzinger, Dennis, Simon Heuschkel, and Matthias Drews. 2023.
\newblock {{CICL}}\_{{DMS}} at {{SemEval-2023 Task}} 11: {{Learning With
  Disagreements}} ({{Le-Wi-Di}}).
\newblock In \emph{Proceedings of the 17th {{International Workshop}} on
  {{Semantic Evaluation}} ({{SemEval-2023}})}, pages 1030--1036.

\bibitem[{Harju and Mesaros(2023)}]{harju2023}
Harju, Manu and Annamaria Mesaros. 2023.
\newblock Evaluating classification systems against soft labels with fuzzy
  precision and recall.
\newblock In \emph{Proceedings of the 8th Detection and Classification of
  Acoustic Scenes and Events 2023 Workshop ({{DCASE2023}})}, pages 46--50.

\bibitem[{Hu et~al.(2022)Hu, Shen, Wallis, {Allen-Zhu}, Li, Wang, Wang, and
  Chen}]{hu2022}
Hu, Edward~J., Yelong Shen, Phillip Wallis, Zeyuan {Allen-Zhu}, Yuanzhi Li,
  Shean Wang, Lu~Wang, and Weizhu Chen. 2022.
\newblock {{LoRA}}: {{Low-Rank Adaptation}} of {{Large Language Models}}.
\newblock In \emph{The {{Tenth International Conference}} on {{Learning
  Representations}}}.

\bibitem[{Kruse et~al.(2022)Kruse, Mostaghim, Borgelt, Braune, and
  Steinbrecher}]{kruse2022}
Kruse, Rudolf, Sanaz Mostaghim, Christian Borgelt, Christian Braune, and
  Matthias Steinbrecher. 2022.
\newblock Introduction to {{Fuzzy Sets}} and {{Fuzzy Logics}}.
\newblock In \emph{Computational {{Intelligence}}: {{A Methodological
  Introduction}}}. Springer, pages 373--405.

\bibitem[{Kurniawan et~al.(2024)Kurniawan, Mistica, Baldwin, and
  Lau}]{kurniawan2024}
Kurniawan, Kemal, Meladel Mistica, Timothy Baldwin, and Jey~Han Lau. 2024.
\newblock To {{Aggregate}} or {{Not}} to {{Aggregate}}. {{That}} is the
  {{Question}}: {{A Case Study}} on {{Annotation Subjectivity}} in {{Span
  Prediction}}.
\newblock In \emph{Proceedings of the 14th {{Workshop}} on {{Computational
  Approaches}} to {{Subjectivity}}, {{Sentiment}}, \& {{Social Media
  Analysis}}}, pages 362--368.

\bibitem[{Lee, An, and Thorne(2023)}]{lee2023}
Lee, Noah, Na~Min An, and James Thorne. 2023.
\newblock Can {{Large Language Models Capture Dissenting Human Voices}}?
\newblock In \emph{Proceedings of the 2023 {{Conference}} on {{Empirical
  Methods}} in {{Natural Language Processing}}}, pages 4569--4585.

\bibitem[{Leonardelli et~al.(2023)Leonardelli, Abercrombie, Almanea, Basile,
  Fornaciari, Plank, Rieser, Uma, and Poesio}]{leonardelli2023}
Leonardelli, Elisa, Gavin Abercrombie, Dina Almanea, Valerio Basile, Tommaso
  Fornaciari, Barbara Plank, Verena Rieser, Alexandra Uma, and Massimo Poesio.
  2023.
\newblock {{SemEval-2023 Task}} 11: {{Learning}} with {{Disagreements}}
  ({{LeWiDi}}).
\newblock In \emph{Proceedings of the 17th {{International Workshop}} on
  {{Semantic Evaluation}} ({{SemEval-2023}})}, pages 2304--2318.

\bibitem[{Leonardelli et~al.(2021)Leonardelli, Menini, Palmero~Aprosio,
  Guerini, and Tonelli}]{leonardelli2021}
Leonardelli, Elisa, Stefano Menini, Alessio Palmero~Aprosio, Marco Guerini, and
  Sara Tonelli. 2021.
\newblock Agreeing to {{Disagree}}: {{Annotating Offensive Language Datasets}}
  with {{Annotators}}' {{Disagreement}}.
\newblock In \emph{Proceedings of the 2021 {{Conference}} on {{Empirical
  Methods}} in {{Natural Language Processing}}}, pages 10528--10539.

\bibitem[{Lewis and Brown(2001)}]{lewis2001}
Lewis, H.~G. and M.~Brown. 2001.
\newblock A generalized confusion matrix for assessing area estimates from
  remotely sensed data.
\newblock \emph{International Journal of Remote Sensing}, 22(16):3223--3235.

\bibitem[{Lin(1991)}]{lin1991}
Lin, Jianhua. 1991.
\newblock Divergence measures based on the {{Shannon}} entropy.
\newblock \emph{IEEE Transactions on Information Theory}, 37(1):145--151.

\bibitem[{Liu et~al.(2019)Liu, Ott, Goyal, Du, Joshi, Chen, Levy, Lewis,
  Zettlemoyer, and Stoyanov}]{liu2019h}
Liu, Yinhan, Myle Ott, Naman Goyal, Jingfei Du, Mandar Joshi, Danqi Chen, Omer
  Levy, Mike Lewis, Luke Zettlemoyer, and Veselin Stoyanov. 2019.
\newblock {{RoBERTa}}: {{A Robustly Optimized BERT Pretraining Approach}}.
\newblock \emph{CoRR}, cs.CL/1907.11692.

\bibitem[{Maity et~al.(2023)Maity, Kandru, Singh, Aditya~Hari, and
  Varma}]{maity2023}
Maity, Ankita, Pavan Kandru, Bhavyajeet Singh, Kancharla Aditya~Hari, and
  Vasudeva Varma. 2023.
\newblock {{IREL}} at {{SemEval-2023 Task}} 11: {{User Conditioned Modelling}}
  for {{Toxicity Detection}} in {{Subjective Tasks}}.
\newblock In \emph{Proceedings of the 17th {{International Workshop}} on
  {{Semantic Evaluation}} ({{SemEval-2023}})}, pages 2133--2136.

\bibitem[{Negahban, Oh, and Shah(2012)}]{negahban2012}
Negahban, Sahand, Sewoong Oh, and Devavrat Shah. 2012.
\newblock Iterative ranking from pair-wise comparisons.
\newblock In \emph{Advances in {{Neural Information Processing Systems}}},
  volume~25.

\bibitem[{Nie, Zhou, and Bansal(2020)}]{nie2020}
Nie, Yixin, Xiang Zhou, and Mohit Bansal. 2020.
\newblock What can we learn from collective human opinions on natural language
  inference data?
\newblock In \emph{Proceedings of the 2020 Conference on Empirical Methods in
  Natural Language Processing ({{EMNLP}})}, pages 9131--9143.

\bibitem[{Pavlick and Kwiatkowski(2019)}]{pavlick2019}
Pavlick, Ellie and Tom Kwiatkowski. 2019.
\newblock Inherent {{Disagreements}} in {{Human Textual Inferences}}.
\newblock \emph{Transactions of the Association for Computational Linguistics},
  7:677--694.

\bibitem[{Peterson et~al.(2019)Peterson, Battleday, Griffiths, and
  Russakovsky}]{peterson2019}
Peterson, Joshua, Ruairidh Battleday, Thomas Griffiths, and Olga Russakovsky.
  2019.
\newblock Human {{Uncertainty Makes Classification More Robust}}.
\newblock In \emph{2019 {{IEEE}}/{{CVF International Conference}} on {{Computer
  Vision}} ({{ICCV}})}, pages 9616--9625.

\bibitem[{Plank(2022)}]{plank2022}
Plank, Barbara. 2022.
\newblock The ``{{Problem}}'' of {{Human Label Variation}}: {{On Ground Truth}}
  in {{Data}}, {{Modeling}} and {{Evaluation}}.
\newblock In \emph{Proceedings of the 2022 {{Conference}} on {{Empirical
  Methods}} in {{Natural Language Processing}}}, pages 10671--10682.

\bibitem[{Pontius and Cheuk(2006)}]{pontius2006}
Pontius, R.~G. and M.~L. Cheuk. 2006.
\newblock A generalized cross-tabulation matrix to compare soft-classified maps
  at multiple resolutions.
\newblock \emph{International Journal of Geographical Information Science},
  20(1):1--30.

\bibitem[{Pontius and Connors(2006)}]{pontius2006a}
Pontius, R.~G. and John Connors. 2006.
\newblock Expanding the conceptual, mathematical, and practical methods for map
  comparison.
\newblock In \emph{Proceedings of the {{Meeting}} of {{Spatial Accuracy}}},
  pages 64--79.

\bibitem[{Rizzi et~al.(2024)Rizzi, Leonardelli, Poesio, Uma, Pavlovic, Paun,
  Rosso, and Fersini}]{rizzi2024}
Rizzi, Giulia, Elisa Leonardelli, Massimo Poesio, Alexandra Uma, Maja Pavlovic,
  Silviu Paun, Paolo Rosso, and Elisabetta Fersini. 2024.
\newblock Soft metrics for evaluation with disagreements: An assessment.
\newblock In \emph{Proceedings of the 3rd {{Workshop}} on {{Perspectivist
  Approaches}} to {{NLP}} ({{NLPerspectives}}) @ {{LREC-COLING}} 2024}, pages
  84--94.

\bibitem[{{Rodr{\'i}guez-Barroso} et~al.(2024){Rodr{\'i}guez-Barroso},
  C{\'a}mara, Collados, Luz{\'o}n, and Herrera}]{rodriguez-barroso2024}
{Rodr{\'i}guez-Barroso}, Nuria, Eugenio~Mart{\'i}nez C{\'a}mara, Jose~Camacho
  Collados, M.~Victoria Luz{\'o}n, and Francisco Herrera. 2024.
\newblock Federated {{Learning}} for {{Exploiting Annotators}}'
  {{Disagreements}} in {{Natural Language Processing}}.
\newblock \emph{Transactions of the Association for Computational Linguistics},
  12:630--648.

\bibitem[{Sheng, Provost, and Ipeirotis(2008)}]{sheng2008}
Sheng, Victor~S., Foster Provost, and Panagiotis~G. Ipeirotis. 2008.
\newblock Get another label? improving data quality and data mining using
  multiple, noisy labelers.
\newblock In \emph{Proceedings of the 14th {{ACM SIGKDD}} International
  Conference on Knowledge Discovery and Data Mining}, pages 614--622.

\bibitem[{{Silv{\'a}n-C{\'a}rdenas} and Wang(2008)}]{silvan-cardenas2008}
{Silv{\'a}n-C{\'a}rdenas}, J.~L. and L.~Wang. 2008.
\newblock Sub-pixel confusion--uncertainty matrix for assessing soft
  classifications.
\newblock \emph{Remote Sensing of Environment}, 112(3):1081--1095.

\bibitem[{Sullivan, Yasin, and Jacobs(2023)}]{sullivan2023}
Sullivan, Michael, Mohammed Yasin, and Cassandra~L. Jacobs. 2023.
\newblock University at {{Buffalo}} at {{SemEval-2023 Task}} 11:
  {{MASDA}}--{{Modelling Annotator Sensibilities}} through {{DisAggregation}}.
\newblock In \emph{Proceedings of the 17th {{International Workshop}} on
  {{Semantic Evaluation}} ({{SemEval-2023}})}, pages 978--985.

\bibitem[{Trager et~al.(2022)Trager, Ziabari, Davani, Golazizian,
  {Karimi-Malekabadi}, Omrani, Li, Kennedy, Reimer, Reyes, Cheng, Wei,
  Merrifield, Khosravi, Alvarez, and Dehghani}]{trager2022a}
Trager, Jackson, Alireza~S. Ziabari, Aida~Mostafazadeh Davani, Preni
  Golazizian, Farzan {Karimi-Malekabadi}, Ali Omrani, Zhihe Li, Brendan
  Kennedy, Nils~Karl Reimer, Melissa Reyes, Kelsey Cheng, Mellow Wei, Christina
  Merrifield, Arta Khosravi, Evans Alvarez, and Morteza Dehghani. 2022.
\newblock The {{Moral Foundations Reddit Corpus}}.
\newblock \emph{CoRR}, cs.CL/2208.05545.

\bibitem[{Uma et~al.(2020)Uma, Fornaciari, Hovy, Paun, Plank, and
  Poesio}]{uma2020}
Uma, Alexandra, Tommaso Fornaciari, Dirk Hovy, Silviu Paun, Barbara Plank, and
  Massimo Poesio. 2020.
\newblock A {{Case}} for {{Soft Loss Functions}}.
\newblock \emph{Proceedings of the AAAI Conference on Human Computation and
  Crowdsourcing}, 8(1):173--177.

\bibitem[{Uma et~al.(2021)Uma, Fornaciari, Hovy, Paun, Plank, and
  Poesio}]{uma2021}
Uma, Alexandra~N., Tommaso Fornaciari, Dirk Hovy, Silviu Paun, Barbara Plank,
  and Massimo Poesio. 2021.
\newblock Learning from {{Disagreement}}: {{A Survey}}.
\newblock \emph{Journal of Artificial Intelligence Research}, 72:1385--1470.

\bibitem[{Vitsakis et~al.(2023)Vitsakis, Parekh, Dinkar, Abercrombie, Konstas,
  and Rieser}]{vitsakis2023}
Vitsakis, Nikolas, Amit Parekh, Tanvi Dinkar, Gavin Abercrombie, Ioannis
  Konstas, and Verena Rieser. 2023.
\newblock {{iLab}} at {{SemEval-2023 Task}} 11 {{Le-Wi-Di}}: {{Modelling
  Disagreement}} or {{Modelling Perspectives}}?
\newblock In \emph{Proceedings of the 17th {{International Workshop}} on
  {{Semantic Evaluation}} ({{SemEval-2023}})}, pages 1660--1669.

\bibitem[{Wan and {Badillo-Urquiola}(2023)}]{wan2023}
Wan, Ruyuan and Karla {Badillo-Urquiola}. 2023.
\newblock Dragonfly\_captain at {{SemEval-2023 Task}} 11: {{Unpacking
  Disagreement}} with {{Investigation}} of {{Annotator Demographics}} and
  {{Task Difficulty}}.
\newblock In \emph{Proceedings of the 17th {{International Workshop}} on
  {{Semantic Evaluation}} ({{SemEval-2023}})}, pages 1978--1982.

\bibitem[{Zhang et~al.(2023)Zhang, Malkov, Florez, Park, McWilliams, Han, and
  {El-Kishky}}]{zhang2023a}
Zhang, Xinyang, Yury Malkov, Omar Florez, Serim Park, Brian McWilliams, Jiawei
  Han, and Ahmed {El-Kishky}. 2023.
\newblock {{TwHIN-BERT}}: {{A Socially-Enriched Pre-trained Language Model}}
  for {{Multilingual Tweet Representations}} at {{Twitter}}.
\newblock In \emph{Proceedings of the 29th {{ACM SIGKDD Conference}} on
  {{Knowledge Discovery}} and {{Data Mining}}}, pages 5597--5607.

\end{thebibliography}

\end{document}